\pdfoutput=1
\documentclass[11pt]{article}

\usepackage{acl}
\usepackage{times}
\usepackage{latexsym}
\usepackage{todonotes}[disable]
\usepackage{booktabs}
\usepackage{amsfonts}
\usepackage{xcolor}
\usepackage{textcomp}
\usepackage{multirow}
\usepackage{multicol}
\usepackage[normalem]{ulem}
\usepackage[many]{tcolorbox}
\usepackage[T1]{fontenc}
\usepackage[utf8]{inputenc}
\usepackage{mathrsfs}
\usepackage{algorithm,algpseudocode}
\usepackage{amsmath}
\usepackage{arydshln}
\usepackage{microtype}

\usepackage{inconsolata}

\usepackage{graphicx}
\usepackage{subcaption}
\usepackage[export]{adjustbox}
\newcommand{\cil}{{\em CIL}}
\newcommand{\pare}{{\em PARE}}
\newcommand{\hydra}{{\sc Hydre}}
\newcommand{\parex}{{\em PARE-X}}
\newcommand{\cilx}{{\em CIL-X}}

\title{Combining Distantly Supervised Models with In Context Learning for Monolingual and Cross-Lingual Relation Extraction}

\author{Vipul Rathore \hskip 1em  
Malik Hammad Faisal  \hskip 1em
Parag Singla \hskip 1em Mausam \\
        Indian Institute of Technology \\
  New Delhi, India \\  
   \texttt\{vipul.rathore, cs1210559, parags, mausam\}@cse.iitd.ac.in}

\newtcolorbox{examplebox}{
    colback=blue!5!white,
    colframe=blue!50!black,
    fonttitle=\bfseries,
    title=ICL Exemplars (\hydra),
    sharp corners
}
\newtcolorbox{querybox}{
    colback=green!5!white,
    colframe=green!50!black,
    fonttitle=\bfseries,
    title=Query:,
    sharp corners
}
\newtcolorbox{outputbox}{
    colback=orange!5!white,
    colframe=orange!50!black,
    fonttitle=\bfseries,
    title=Outputs:,
    sharp corners
}
\newcommand{\correct}[1]{\textcolor{green!70!black}{\textbf{#1}}}
\newcommand{\wrong}[1]{\textcolor{red!70!black}{\textbf{#1}}}

\begin{document}
\maketitle
\begin{abstract}

Distantly Supervised Relation Extraction (DSRE) remains a long-standing challenge in NLP, where models must learn from noisy bag-level annotations while making sentence-level predictions. While existing state-of-the-art (SoTA) DSRE models rely on task-specific training, their integration with in-context learning (ICL) using large language models (LLMs) remains underexplored. A key challenge is that the LLM may not learn relation semantics correctly, due to noisy annotation. 

In response, we propose \hydra{} -- \textbf{\underline{HY}}brid \textbf{\underline{D}}istantly Supervised \textbf{\underline{R}}elation \textbf{\underline{E}}xtraction framework. It first uses a trained DSRE model to identify the top-$k$ candidate relations for a given test sentence, then uses a novel dynamic exemplar retrieval strategy that extracts reliable, sentence-level exemplars from training data, which are then provided in LLM prompt for outputting the final relation(s).
%and then leverages an LLM with ICL to select the most suitable relation(s). To overcome the challenge of label noise in distant supervision, we introduce a dynamic exemplar retrieval strategy that extracts reliable, sentence-level exemplars from noisy bag-level training data.
%\newline
We further extend \hydra{} to cross-lingual settings for RE in low-resource languages. Using available English DSRE training data, we evaluate all methods on English as well as a newly curated benchmark covering four diverse low-resource Indic languages -- Oriya, Santali, Manipuri, and Tulu. \hydra\ achieves up to 20 F1 point gains in English and, on average, 17 F1 points on Indic languages over prior SoTA DSRE models. Detailed ablations exhibit \hydra{}'s efficacy compared to other prompting strategies.
    
\end{abstract}

\section{Introduction}
% \textit{Relation extraction} (RE) has been a widely recognized problem in the wider \textit{Information Extraction } (IE) community. Given a sentence $s$ mentioning 2 entities - \textit{head} entity $h$ and \textit{tail} entity $t$, the task is to predict all relations $\hat{r}$s $\in \mathcal{R}$, where $\mathcal{R}$ is a given predefined ontology of relations. 

% This requires manually curating a training data which is costly. To mitigate this, \citet{mintz2009distant} introduced \textit{Distantly Supervised Relation Extraction} (DSRE), which labels the training data automatically by aligning the Wikipedia sentences with an external knowledge base such as Wikidata. This creates a training data in which each instance is a \textit{bag} $B(e_1,e_2)$ of sentences mentioning entity pairs $e_1$ and $e_2$ with relations labels $R(e_1,e_2)$ i.e. set of all relations between $e_1$ and $e_2$ in a KB. This is different from supervised setting as sentence-level annotations are not available. This is challenging since a model must be able to learn from bag-level training data, but must be able to extract relation(s) from a query sentence at inference.
Relation Extraction (RE) is a core task in Information Extraction (IE) that aims to identify semantic relations between entity mentions in text. Given a sentence $s$ containing a head entity $e_1$ and a tail entity $e_2$, the goal is to predict the set of relations $(\hat{r} \subset \mathcal{R})$ expressed between them, where $\mathcal{R}$ is a predefined ontology of relations. RE plays a crucial role in downstream applications such as knowledge base construction, and question answering. Supervised RE methods depend on sentence-level annotations, which are costly and time-consuming to obtain at scale \cite{zhang-etal-2017-position}.

% To mitigate this bottleneck, \citet{mintz2009distant} introduced Distantly Supervised Relation Extraction (DSRE), a weakly supervised paradigm that automatically aligns text with structured knowledge base (KB) triples (e.g., from Wikidata). This alignment generates training data in the form of bags 
% $B(e_1,e_2)$, where each bag contains sentences mentioning a given entity pair ($e_1$, $e_2$) -- a bag is labeled with all known relations $R(e_1,e_2)$ between them in the KB. Unlike traditional supervised setting, DSRE provides only noisy bag-level supervision, while requiring sentence-level predictions during inference—posing unique learning challenges.
Distantly Supervised Relation Extraction (DSRE) \cite{mintz2009distant} alleviates this challenge by aligning text with knowledge base (KB) triples to generate weakly labeled training data. Specifically, DSRE groups all sentences mentioning an entity pair ($e_1$, $e_2$) into a bag $B(e_1,e_2)$, which is labeled with all relations $R(e_1,e_2)$ known between ($e_1$, $e_2$) in the KB.  Although the supervision is weak and bag-level, inference is typically performed at the sentence level (micro-reading \cite{mitchell-iswc09}).
This mismatch between training and inference granularity -- coupled with noisy training labels -- often causes state-of-the-art (SoTA) DSRE models to confuse fine-grained relation types, such as \textit{Nationality} vs. \textit{Place\_of\_Birth} or \textit{Founder} vs. \textit{CEO}, thereby limiting their overall performance.

Existing DSRE methods \cite{chen-etal-2021-cil, rathore2022pare, jian2024distantly} primarily rely on task-specific fine-tuning of moderate-sized language models such as BERT \cite{devlin-etal-2019-bert}  and RoBERTa \cite{liu2019roberta}. However, the potential of Large Language Models (LLMs) for this task remains largely unexplored.  Modern LLMs excel at in-context learning (ICL), where the model performs reasoning by conditioning on a few task-specific exemplars. Yet, directly applying LLMs to DSRE is non-trivial: noisy distant supervision degrades exemplar quality, and the lack of clean, sentence-level exemplars undermines effective ICL. Consequently, prior works in DSRE either ignore LLMs altogether or fail to exploit their reasoning capabilities effectively \cite{zhao2024comprehensive}.

In this work, we propose \hydra: a \underline{\textbf{H}}\underline{\textbf{Y}}brid \underline{\textbf{D}}S\underline{\textbf{R}}\underline{\textbf{E}} framework that combines the high-recall candidate label selection of fine-tuned DSRE models with the reasoning capabilities of LLMs. Given a query sentence, a fine-tuned DSRE model %-- such as PARE \cite{rathore2022pare} or CIL \cite{chen-etal-2021-cil} -- %
first identifies a candidate relation set by filtering out the obvious negatives. These candidates are then passed to an LLM for disambiguation, guided by carefully selected in-context exemplars. To construct these exemplars, we retrieve relevant bags from the DSRE training corpus using a joint scoring function that blends model confidence and semantic similarity to the query. From each selected bag, we extract the most representative sentence to form a dynamic, relation-specific prompt, guiding the LLM to accurately \textit{judge} the correct relation(s).

We further extend \hydra\ to the cross-lingual setting, focusing on low-resource languages — a largely underexplored area in DSRE. To facilitate this, we construct a new multilingual benchmark covering four low-resource Indic languages: Oriya, Santhali, Manipuri, and Tulu. Given the limited representation of these languages in LLM pretraining corpora \cite{singh-etal-2024-indicgenbench, nag-etal-2025-efficient}, they pose an interesting challenge for evaluating cross-lingual RE in the context of latest LLMs. 

We evaluate \hydra\ under three transfer settings - (1) \textit{English-only-data}, where no target language data is used; (2) \textit{Translate-train}, where English DSRE training data is translated to target language; and (3) \textit{Translate-test}, where test queries in the target language are translated to English.

% \sout{
% For translate-train, we enhance the multilingual in-context learning (ICL) capabilities of small language models (SLMs) like LLaMA 3.1 via LoRA-based fine-tuning on translated DSRE data, and adapt English retrieval strategies using multilingual counterparts of monolingual DSRE models.

% For translate-test, we apply the English pipeline directly to translated queries.

% In the English-only setting, we use off-the-shelf cross-lingual retrievers like BGE-M3 \cite{bge-m3} to select relevant English exemplars.
% }

Experiments with both open-source and proprietary LLMs show that \hydra\ consistently outperforms prior DSRE baselines and naive LLM prompting strategies. Our exemplar retrieval strategy proves robust across both monolingual and cross-lingual setups. Ablation analyses further reveal that removing retrieval components can degrade performance by up to 7 micro-F1 points.

In summary, our key contributions are as follows.
(1) We present the first systematic integration of LLMs via In-Context Learning (ICL) into DSRE inference, achieving significant gains over both fine-tuned and prompting-only baselines. 
(2) We propose a novel retrieval strategy that combines model confidence and semantic similarity to select high-quality ICL exemplars for LLM-based relation disambiguation. 
(3) We curate and release gold-standard evaluation datasets for relation extraction for four typologically diverse low-resource Indic languages. 
(4) We propose effective cross-lingual strategies for adapting \hydra\ to low-resource languages, demonstrating robustness across multiple transfer scenarios. 
We commit to releasing our code and data to facilitate further research in  multilingual DSRE.

\section{Related Work} 

\paragraph{Distantly Supervised Relation Extraction:} DSRE \cite{mintz2009distant} aligns KB triples with text to create bag-level supervision, where labels apply at the bag rather than sentence level. Neural DSRE models typically adopt the multi-instance multi-label (MIML) framework \cite{surdeanu-etal-2012-multi}. Earlier works encoded sentences in a bag using piecewise CNNs \cite{zeng2015distant} or graph CNNs \cite{vashishth-etal-2018-reside}, while recent models employ pre-trained transformers. PARE \cite{rathore2022pare} encodes a bag by treating all bag sentences as a passage, whereas CIL \cite{chen-etal-2021-cil} uses intra-bag attention and contrastive learning. HiCLRE \cite{li2022hiclre} introduces hierarchical contrastive learning, and HFMRE \cite{li2023hfmre} employs Huffman-tree structures to denoise bags. These advances improve bag-level reasoning but still struggle with fine-grained sentence-level inference.

% \textbf{LLMs for DSRE and Data Relabeling:}
% Recently, \citet{jian2024distantly} applied LLMs for relabeling DSRE training sets, followed by finetuning smaller LMs such as BERT. While effective for denoising data, this pipeline requires costly relabeling of the entire corpus and does not leverage LLMs for inference. In contrast, we seek to directly apply LLMs to refine predictions at inference by designing effective ICL example retrieval strategies. Our contributions thus complement \citet{jian2024distantly}’s relabeling approach and could be integrated with it in future work.

\paragraph{Multilingual DSRE:}
% While multilingual supervised Relation Extraction (RE) has seen notable progress in recent years \cite{ni2020cross, nag2021data}, research on multilingual DSRE remains scarce. \citet{bhartiya2021dis} introduced DisRex, a distantly supervised dataset covering four European languages. However, it relies solely on bag-level labels without manually annotated evaluation data, limiting its utility for fine-grained assessment. Moreover, DisRex does not include any low-resource languages, making it inadequate for studying DSRE in the low-resource multilingual settings that our work targets.
Research in multilingual RE has progressed \cite{ni2020cross, nag2021data}, but multilingual DSRE is scarce. DisRex \cite{bhartiya2021dis} introduced a dataset across four European languages, though without manual sentence-level evaluation. No prior work targets typologically diverse low-resource Indic languages, which motivates our benchmark contribution.

% \paragraph{LLMs for supervised RE:} \citet{wadhwa2023revisiting} evaluate the few-shot performance of GPT-3 and show it to be competitive with fully supervised SoTA models. They further elicit chain-of-thought (CoT) reasoning from GPT-3 \cite{NEURIPS2020_1457c0d6} and fine-tune Flan-T5 \cite{chung2024scaling}—a small instruction-tuned language model (SLM)—on CoT-augmented data, achieving new SOTA results on supervised RE benchmarks. However, their approach relies on clean instance-level supervision for generating CoT, limiting its applicability to DSRE, which provides only noisy, bag-level supervision.

% \paragraph{LLMs for DSRE and Data Relabeling:}
% \citet{jian2024distantly} employ LLMs to relabel noisy DSRE training data, followed by supervised finetuning of smaller LMs such as BERT. While effective for denoising, their approach incurs significant computational cost and does not utilize LLMs for inference. In contrast, our approach leverages LLMs directly at inference through a lightweight in-context learning (ICL) framework guided by retrieved exemplars. Our method is thus complementary and potentially integrable with relabeling-based approaches as a future work.

\paragraph{LLMs for Relation Extraction:} Large language models (LLMs) have recently been explored for RE primarily in supervised settings. \citet{wadhwa2023revisiting} demonstrated that few-shot prompting with GPT-3 can rival fully supervised baselines, particularly when enhanced with chain-of-thought (CoT) reasoning. Other works employ zero-shot prompting with relation label definitions \cite{zhou-etal-2024-grasping} or leverage LLMs to denoise distantly supervised training data \cite{li-etal-2023-semi, jian2024distantly}. However, these approaches either assume access to clean, labeled supervision or utilize LLMs solely for data relabeling purposes. 
In contrast, \hydra\ takes a complementary view by employing LLMs directly at test time, leveraging them as reasoning engines rather than annotation tools.

\paragraph{Diversity-aware exemplar selection:}
A common in-context exemplar selection strategy is to retrieve the top-$K$ semantically most similar exemplars to the query, but this often results in redundancy amongst the selected exemplars. To address this, \citet{10.1145/3726302.3730194} propose maximal marginal relevance (MMR), which balances similarity with diversity, while \citet{wang-etal-2024-effective} employ determinantal point processes (DPPs) to encourage diversity through submodular modeling objectives. These methods highlight the importance of complementing similarity with diversity in exemplar selection, a principle that we extend within our hybrid DSRE–LLM framework.

\paragraph{LLM-as-Judge and Hybrid Models:}
Recent work explores LLMs as evaluators (“judges”) for ranking candidate outputs \cite{zheng2023judging, Bavaresco2024JUDGE_BENCH}. Our approach adapts can be seen as an instance, where a DSRE model provides high-recall candidate relations and their exemplars, and the LLM judges among them. Related hybrid paradigms exist in other NLP tasks \cite{rathore-etal-2024-ssp} such as sequence labeling, but not in DSRE.
%ours is the first systematic integration of LLM-as-judge into DSRE inference. 

\begin{figure}[ht]
\centering
\includegraphics[width=0.95\columnwidth, height=5.5cm]{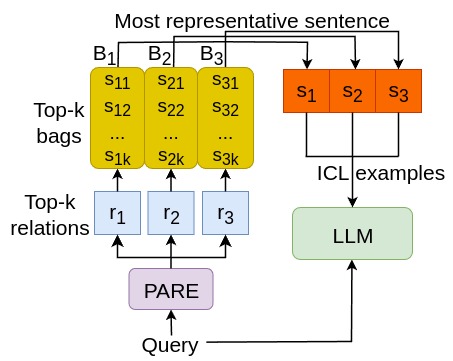}
    \caption{\hydra\ overview (shown for k=3): (a) Select top-k candidate labels using PARE confidence, (b) Select best bag for each candidate using combined semantic similarity and label confidence (aggregated over bag sentences), and (c) Select best sentence per bag using aggregate (over bag labels) confidence scoring.}
    \label{fig:pipeline}
\end{figure}
\section{\hydra: \textbf{\underline{HY}}brid \underline{D}istantly supervised \underline{R}elation \underline{E}xtraction}
%\subsection{Motivation for a Hybrid Framework}
DSRE models are trained on bags $B(e_1,e_2)$, each containing multiple sentences for an entity pair $(e_1,e_2)$ and annotated with a set of relations $R(e_1,e_2)$. Consequently, such models often generalize poorly to sentence-level queries \cite{gao2021manual, chen-etal-2021-cil}. However, our preliminary analysis (Table \ref{table:recall-5}) reveals that SoTA DSRE models such as \pare\ exhibit high Recall@$k$ ($\approx$ 85\% for English) even for small values of $k$ (E.g. $k$=5).

This observation motivates a two-stage hybrid approach: use a trained DSRE model to obtain a high-recall candidate relation set, and employ an LLM as a judge to select the final relation(s). We seek to achieve this by providing carefully selected exemplars for these candidate relations to LLM. Effective use of in-context learning (ICL), however, requires clean, sentence-level exemplars—whereas DSRE training examples are noisy and bag-level. To bridge this gap, \hydra\ employs a novel three-stage exemplar selection process described below. 
%\subsection{Three-Stage Exemplar Selection}
%Our exemplar selection pipeline (Fig. \ref{fig:pipeline}) consists of three stages:
\paragraph{Stage 1: Candidate Relation Selection.}  
A trained DSRE model (e.g., \pare) assigns confidence scores $f_{\text{PARE}}(q, r)$ to each relation $r$ for a query sentence $q$. The top-$k$ relations with the highest scores form the candidate relation set $\mathcal{R}'$.
%, which are passed onto Stage 2 for further processing.
% We use a trained DSRE model to compute scores $f(q, r)$ for all relations $r$. The top-$k$ relations with the highest scores form the candidate set $\mathcal{R}'$. Our DSRE model achieves decent Recall@$k$ even for smaller values of $k$ (Table~\ref{table:recall-5}), reducing prompt length and improving LLM focus. We set $k=5$ in all experiments.

\paragraph{Stage 2: Bag Selection.}  
% Distant supervision models are typically trained on bag-level data, where each bag contains multiple sentences mentioning an entity pair, and a set of relation labels. However, such models often generalize poorly when applied to a sentence-level query at inference. To mitigate this, we divide exemplar selection into two stages. In the first stage, for each of the top-$k$ candidate relations, we identify the most relevant bag $B_j$ from the subset of bags for which the relation $r$ is included in their label set. This ensures that the selected bag is actually annotated (via distant supervision) with the relation we are trying to exemplify. We compute a combined score consisting of the semantic similarity between the query $q$ and the bag $B_j$, and the DSRE model score $f(B_j, r)$ — which evaluates how confidently the model predicts relation $r$ for the concatenated bag. This encourages the selection of bags that are both semantically similar to the query and informative w.r.t. the target relation.
% We aim to select semantically relevant bag for each of the candidate relations $r$. For this, we first consider only those bags $\mathcal{B}_r$ having relation $r$ in their relation set. Then we score each bag $B_j \in \mathcal{B}_r$ using a combined score of the semantic similarity $\mathrm{sim}(q, B_j)$ and the DSRE model confidence $f(B_j, r)$, computed using \red{max} pooling over sentence scores in the bag. The bag $B_r$ having the highest score is chosen for $r$.
For each candidate relation $r \in \mathcal{R}'$, we identify the most relevant bag $B_r$ from the subset $\mathcal{B}_r$ of bags annotated with relation $r$ as one of it's relations. Each bag $B_j \in \mathcal{B}_r$ is scored using a weighted combination of ($i$) semantic similarity $\mathrm{sim}(q, B_j)$ and ($ii$) DSRE model confidence $f(B_j, r)$, computed using \textit{max}-pooling over its sentence scores. The highest-scoring bag is chosen as $B_r$ for relation $r$. Essentially, this chosen bag has sentences that are strongly representative of $r$, while being similar to the test sentence $q$.

\paragraph{Stage 3: Sentence Selection.}  
% In real-world use cases, each query corresponds to a single sentence, not a group of sentences as in distant supervision. Hence, to ensure compatibility between the query and in-context learning (ICL) exemplars, we further extract a single \textit{representative} sentence from each selected bag. We first identify all sentences in the bag which maximize the count of relations $r_a$ (where $r_a \in$ the bag’s label set) such that $f(s, r_a) > t$  i.e., relations whose confidence is more than a threshold. Among these, we choose the sentence that has the highest aggregated confidence $\sum_{r_a} f(s, r_a)$. This ensures that the selected sentence is both confident and representative of the relation labels of selected bag. We justify our design choice for using the aggregate confidence for sentence selection using additional ablations shown in Appendix \ref{subsec: ablat3}. \\
% The full selection process is detailed in Algo. ~\ref{alg:selecting_exemplars}.
% From each selected bag $B_r$, we seek to choose a single representative sentence that expresses most of the bag labels. For each $s \in B_r$, we first compute the coverage $c(s) = \sum_{r_a \in \text{labels}(B_r)} \mathbb{I}[f(s, r_a) > t]$ with confidence threshold $t$. Among sentences with maximum coverage, we select the one with highest aggregate confidence $\sum_{r_a} f(s, r_a)$. This would try to enforce selecting sentences covering most of the bag labels with a decently high confidence. \newline
% Exemplars are then ordered by ascending $f(q,r)$, placing more informative examples closer to the query in the prompt.
From each selected bag $B_r$, we extract a single sentence $s_r$ that best captures the relation(s) expressed in the bag. For each sentence $s \in B_r$, we compute its coverage
$c(s) = \sum_{r_a \in \text{labels}(B_r)} \mathbb{I}[f(s, r_a) > t]$,
where $t$ is a confidence threshold. Among sentences with maximum coverage, we select $s_r$ as the one with the highest aggregate confidence $\sum_{r_a} f(s, r_a)$. This encourages selection of sentences that express the largest number of valid bag relations with strong model confidence.  In the prompt we place the selected sentences $s_r$ along with their corresponding bag labels $\text{\textit{labels}}(B_{r})$. 

Our stage 3 offers three key benefits. First, it aids the \textit{multi-label} nature of the problem, since co-occuring relations (which may be present in $B_r$, but not in $\mathcal{R}'$) are also added to the prompt. This may not have been possible if the method tried to, instead, search for sentences that express only the specific relation. Second, selecting a sentence that has high label confidence over many relations promotes diversity (coverage) of relevant labels in the prompt -- this may not occur if only semantic similarity with test sentence were used for selection. Finally, scoring sentences using aggregate confidence over all bag labels helps surface sentences with stronger overall evidence, not just noisy single-label confidence (as justified via ablations shown in subsec. \ref{subsec: agg_score}).

Once selected, exemplars are ordered by ascending $f(q,r)$, placing more relevant candidates closer to the query in the prompt. The full selection process is described in Algo. ~\ref{alg:selecting_exemplars} (appendix).

\paragraph{Prompting and Parsing. }
Each prompt comprises ($i$) task instruction, ($ii$) candidate relation ontology along with their definitions, ($iii$) ICL exemplars, and ($iv$) the query. The exact prompt template is shown in subsec. \ref{subsec:prompt}.  The output is parsed using string matching to obtain predicted labels, including ``NA". In case no valid match is found, we label it as ``NA".

\section{Dataset Curation}

%\subsection{Cross-lingual evaluation sets}
\label{subsec: data}
% To extend \hydra\ framework to low-resource languages (LRLs), we construct evaluation datasets in four Indic LRLs: Oriya, Manipuri, Santhali and Tulu. We begin by creating an English test split from NYT-10m \cite{gao2021manual} testset using stratified sampling (details in subsec. \ref{subsec: testdata}). 

% We then translate this English test set into each target language using constrained translation via \textit{CODEC} \cite{le2024constrained}, which jointly performs translation and \{head, tail\} entity projection. All translations are manually verified and corrected by a native speaker for each language. To assess reliability, a second annotator evaluates inter-annotator agreement on 100 randomly sampled sentences in each language. We observe agreement exceeding 90\% across all four languages for both translations and entity projections. Complete details follow in appendix subsec. \ref{subsec:indictest}.
To evaluate \hydra\ in low-resource settings, we construct gold-standard test sets for four Indic languages -- Oriya, Santali, Manipuri, and Tulu. We choose these languages because they are an orthographically diverse set of languages that have limited presence on the Web (3K-25K Wikipedia articles), but have large number of native speakers (over 1 million each). This makes them a good challenge set for testing modern NLP systems.

We begin with a stratified subset of the English NYT-10m test split \cite{gao2021manual}, having 538 multi-label queries with a total of 722 labels (incl. 30 ``NA"s), ensuring balanced coverage across relation types. Each English sentence is translated into the target language using CODEC (\textit{constrained decoding} \cite{le2024constrained}) with IndicTrans2 \cite{gala2023indictrans} as the underlying model. CODEC performs joint translation and entity projection to preserve head–tail spans, ensuring that entity mentions remain correctly aligned in the target script. For Tulu, where IndicTrans2 lacks coverage, we use the Google Translate API.

To further enhance entity preservation, we construct synthetic parallel data using entity-span-alignment heuristics (following \citealp{chen2023frustratingly}) and fine-tune IndicTrans2 before translation.
All translations are subsequently verified and corrected by native speakers fluent in reading and typing their respective scripts. Annotators confirmed the correctness of both the full sentence translation as well as the \{head($e1$), tail($e2$)\} entity projections. A second annotator independently re-evaluated 100 randomly sampled sentences per language to measure reliability.  

Across all languages, over 70\% of translations required no correction, while for the corrected ones, their character-level F1 match with original outputs averaged 93\%. Inter-annotator agreement exceeded 90\% for both sentence translations and entity projections (Table~\ref{table:inter-ann}), demonstrating strong consistency and translation quality.  Detailed statistics on dataset sizes, label distributions, and annotation guidelines are provided in Appendix~\ref{subsec:indictest}.

\section{Experiments}
%\subsection{Cross-lingual Settings}
We do two sets of experiments: monolingual (English) and cross-lingual transfer (in the four low-resource Indic languages). For cross-lingual transfer experiments, 
we evaluate \hydra\ across three standard settings commonly used in multilingual NLP.  
Let $X_{\text{train}}$ denote the translated DSRE training corpus from English to target language $X$.  
We define \textit{PARE-X} and \textit{CIL-X} as the target-language counterparts of \textit{PARE} and \textit{CIL}, fine-tuned respectively on $X_{\text{train}}$.  The settings are:

\begin{enumerate}
\itemsep=0em
    \item \textbf{English-only:} Models are trained and prompted exclusively on English DSRE data.
    For English test queries, \hydra\ uses the PARE model’s confidence scores $f_{\text{PARE}}(q, r)$ for candidate ranking and computes semantic similarity using the off-the-shelf encoder \textit{e5-large-v2} \cite{wang2022text}.
    For Indic test queries, since PARE or CIL are not trained on Indic scripts, confidence scores are omitted; \hydra\ instead relies solely on the multilingual encoder \textit{BGE-M3} \cite{bge-m3} to compute cross-lingual semantic similarity.
    \item \textbf{Translate-train:}  
%The English DSRE training corpus is translated into the target language $X$, producing $X_{\text{train}}$.  
\hydra$_{X}$ employs \textit{PARE-X} for confidence estimation and \textit{CIL-X} encoders—trained via contrastive learning to generate high-quality sentence embeddings in $X$ for semantic similarity.

\item \textbf{Translate-test:}  
Test queries in the target language $X$ are translated into English ($X_{\text{test}}$), after which the standard English \hydra\ pipeline is applied directly.
\end{enumerate}

\begin{table}[ht]
\centering
\small{
\begin{tabular}{@{}l|l|l@{}}
  \textbf{Setting} & \textbf{Similarity Model} & \textbf{Confidence Model} \\ 
 \hline
 English-only (En) & e5-large-v2 & PARE \\
 English-only (Indic) & BGE-M3 & -- \\
 Translate-train & CIL-X & PARE-X \\
 Translate-test & e5-large-v2 & PARE \\
\end{tabular}}
\caption{Models used for semantic similarity and confidence estimation across various evaluation settings.}
\label{table:sim-stats}
\end{table}

Table~\ref{table:sim-stats} summarizes the model configurations for each setting.  
In Table~\ref{tab:1}, we denote a variant as \hydra($M$) when using LLM $M$ as the judge, and as \hydra$_X$($M$) when exemplars are retrieved from $X_{\text{train}}$ under the \textit{Translate-train} setting.

\paragraph{Baselines.}  
We compare \hydra\ against the following categories of baselines: %(notations kept consistent with Table ~\ref{tab:1}):
\begin{itemize}
\itemsep=0em
    \item \textbf{Supervised DSRE models:} PARE and CIL for English, and their target-language counterparts PARE-X and CIL-X trained under the \textit{Translate-train} setting.
    \item \textbf{0-shot prompting LLM:} We evaluate both open and proprietary large language models: \textit{Qwen3-235B-A22B}, \textit{GPT-4o}, \textit{Llama3.1-8B}, and its fine-tuned variant \textit{Llama3.1-8B-FT}.  
Here, \textit{Llama3.1-8B-FT} denotes a model fine-tuned using English DSRE data for English experiments, and on $X_{\text{train}}$ for the \textit{Translate-train} setting.   We consider two prompting variants.
\textbf{Direct:} LLM is provided only with the query sentence and the list of candidate relation labels.
\textbf{Ontology-based:} LLM additionally receives the definitions of candidate relations to aid disambiguation.

    \item \textbf{Few-shot prompting LLM:} We further compare against few-shot prompting strategies that differ in exemplar retrieval. 
        \textbf{Random-K} randomly selects $K$ exemplars for each query.
        \textbf{TopK-sim} retrieves the top-$K$ semantically most similar exemplars to the query sentence (similarity-based selection).
        \textbf{LM-MRR} employs a diversity-aware selection strategy using a Maximal Marginal Relevance (MMR) objective \cite{10.1145/3726302.3730194} to balance semantic relevance and diversity. Details are provided in Appendix~\ref{subsubsec: baselines}. 
    \end{itemize}

\paragraph{English Datasets.}  
We use NYT-10m and Wiki-20m datasets \cite{gao2021manual} for English evaluations. Due to space limitations, results and discussion on Wiki-20m are deferred to Appendix  \ref{subsec: wiki}.

\paragraph{Implementation details.}
For obtaining $X_{train}$ and $X_{test}$ in our experiments, we use EasyProject \cite{chen2023frustratingly}, a more lightweight joint translation and entity projection tool as compared to CODEC. %EasyProject is used exclusively for generating large-scale translated training and test data used in our experimental pipeline.
We use LoRA fine-tuning for LLaMA-3.1, updating only adapter and embedding weights while freezing other parameters (App.~\ref{subsec:finetune}). For PARE-X training, we continually pretrain mBERT on $X_{\text{train}}$ before adapter-based fine-tuning.  
All LLMs are run with temperature 0.0, max input tokens 2048, and max generation tokens 256.  

% For semantic similarity, we adopt \textit{e5-large-v2} for English queries, BGE-M3 for Indic queries in the \textit{English-only} setting,\footnote{BGE-M3 is a multilingual encoder capable of aligning cross-lingual representations, making it a suitable fallback when DSRE confidence is unavailable.} and CIL-X encoders in \textit{translate-train}.  
% For constrained translation, we employ CODEC \cite{le2024constrained} with IndicTrans2, except for Tulu where Google Translate API is used. To better preserve entity markers, we construct synthetic parallel data using entity-span-alignment heuristics (following \cite{chen2023frustratingly}) before fine-tuning IndicTrans2. Following resource constraints, CODEC is used primarily for test-set creation, while EasyProject is used elsewhere.  

\paragraph{Evaluation Metrics.}  
We report micro-F1 and macro-F1. Area-under-curve (AUC) is omitted as LLMs cannot yield calibrated confidence scores. Statistical significance is tested using McNemar’s test \cite{mcnemar1947note} for micro-F1.

\begin{table*}[t!]  % Use [t!] for better placement control
% \begin{minipage}{0.45\textwidth}
\small
\centering
%\begin{tabular}{l|c|c|c|c|c|}
%\begin{tabular}{l c c c c}
\begin{tabular}{l c ccc}
\hline
% \textbf{Model} & \textbf{English} & \multicolumn{3}{c}{\textbf{Indic}} \\
\multirow{2}{*}{\textbf{Model}} & \multirow{2}{*}{\textbf{English}} & \multicolumn{3}{c}{\textbf{Indic Languages}} \\
\cmidrule(lr){3-5}
 &  & \textbf{English-only} & \textbf{Translate-train} & \textbf{Translate-test} \\
\hline
%  & \textbf{English} & \textbf{Indic} & & \\
% \midrule
\textit{Supervised} \\
% HICLRE \cite{li2022hiclre} & 31/18 & --  & -- & -- & -- & -- & -- \\
% HFMRE \cite{li2023hfmre} & 32/18 & --  & -- & -- & -- & -- & -- \\
\textit{HFMRE} \cite{li2023hfmre} & 33/18 & -- & 22/13 & 27/16 \\
\textit{HiCLRE} \cite{li2022hiclre} & 31/18 & -- & 20/13 & 25/14 \\
\cil\ \cite{chen-etal-2021-cil} & 43/32 & -- & 26/18 & 34/24  \\
\pare\ \cite{rathore2022pare} & 42/31 & -- & 30/20 & 33/23\\
\hline
\hline
\textit{0-shot (direct)} \\
%Gemma3\text{-}4b &19/17 & 15/13 & 15/13 & 19/16 \\
Qwen3-235B-A22B & 47/39 & 25/21 & 25/21 & 40/34\\
Llama3.1\text{-}8b & 24/21 & 16/12 & 16/12 & 22/20\\
Llama3.1-8B-FT & 55/37 & 26/17 & 26/17 & 47/31\\
GPT-4o & 56/55 & 31/29 & 31/29 & 51/49\\
\midrule
\textit{0-shot (Ontology-based)} \\
%Gemma3\text{-}4b &19/17 & 15/13 & 15/13 & 19/16 \\
Qwen3-235B-A22B & 49/40 & 25/21 & 25/21 & 44/35 \\
Llama3.1\text{-}8b & 31/17 & 22/16 & 22/16 & 29/25 \\
Llama3.1-8B-FT & 60/44 & 24/14 & 38/23 & 49/34 \\
GPT-4o & 56/57 & 33/31 & 32/30 & 54/51 \\
%\bottomrule
% \end{tabular}
% \end{minipage} \hfill
% \begin{minipage}{0.45\textwidth}
% \begin{tabular}{@{}lrrr@{}}
\midrule
\hline
\textit{few-shot (Random)} \\
%TopK-sim(Gemma3-4B) &  &  \\
Random-K(Qwen3-235B-A22B) & 55/55  & 21/19 & 29/25 & 43/35\\
Random-K(Llama3.1-8B) & 30/24 & 19/11 & 11/7 & 27/21 \\
Random-K(Llama3.1-8B-FT) & 55/40 & 24/14 & 25/12 & 46/32 \\
Random-K(GPT-4o) & 56/52 & 31/30 & 38/31 & 48/42 \\
\midrule
\textit{few-shot (Similarity-based)} \\
%TopK-sim(Gemma3-4B) & 16/14  &  \\
TopK-sim(Qwen3-235B-A22B) & 52/49 & 22/18 & 32/26 & 45/38\\
TopK-sim(Llama3.1-8B) & 33/27 & 22/14 & 19/9 & 30/22\\
TopK-sim(Llama3.1-8B-FT) & 55/40 & 27/15 & 29/14 & 46/30 \\
TopK-sim(GPT-4o) & 58/53 & 29/26 & 41/32 & 48/41\\
\midrule
\textit{few-shot (Diversity-based)} \\
%LM-MRR(Gemma3-4B) & 16/14  &  \\
LM-MRR(Qwen3-235B-A22B) & 50/47 & 24/20 & 33/28 & 44/36 \\
LM-MRR(Llama3.1-8B) & 34/26 & 21/12 & 17/8 & 30/22 \\
LM-MRR(Llama3.1-8B-FT) & 56/40 & 26/16 &  28/15 & 47/32 \\
LM-MRR(GPT-4o) & 56/52  &  28/25 & 41/33 & 50/43\\
\midrule
\textit{few-shot (\hydra)} \\
%HYDRE(Gemma3-4B) & 24/19 & 16/14 & 19/14 & 20/16\\
HYDRE(Qwen3-235B-A22B) & \textbf{63*}/\textbf{62} & 29/26 & 38/31 & 54/46\\
HYDRE(Llama3.1-8B) & 52/47 & 30/19 & 35/21 & 44/36\\
HYDRE(Llama3.1-8B-FT) & 61/45 & 35/21 & \textbf{47*}/29 & 51/37\\
HYDRE(GPT-4o) & \textbf{63*}/60 & \textbf{36*}/\textbf{33} & 38/\textbf{34} & \textbf{56*}/\textbf{54}\\
\bottomrule
\end{tabular}
% \end{minipage}
\caption{Results for English and Indic languages (across 3 cross-lingual settings). In each entry, we report micro and macro F1 scores. The Indic results are averaged over 4 languages (language-wise results shown in App. Tables \ref{tab:1_lang}, \ref{tab:2_lang} and \ref{tab:3_lang}). * McNemar's p-value $<10^{-5}$ (valid for micro-F1 comparison).}
\label{tab:1}
\end{table*}

\section{Results \& Analysis} 
% We report results for all three experimental settings in Tables \ref{tab:1}, \ref{tab:2} and \ref{tab:3}. These experiments evaluate the robustness of our few-shot technique in both monolingual and cross-lingual settings, including translate-train (English -> X) and translate-test (X -> English). In Setting I, no data from the target language is used. 
Table \ref{tab:1} reports results on English and four Indic languages across the three cross-lingual settings. 
\paragraph{English results.}
Among supervised DSRE models, \cil\ (43 micro-F1) and \pare\ (42 micro-F1) are strongest.
Zero-shot prompting with GPT-4o already surpasses them (56 F1), and \hydra’s few-shot prompting further lifts GPT-4o to 63 F1.
Smaller open LLMs (Llama 3.1, Qwen 3) gain even more—up to +14 F1—demonstrating that \hydra’s exemplar-guided prompting yields larger benefits for weaker LLMs.
These gains highlight the value of \hydra’s label-aware exemplar selection, which leverages DSRE model confidences to guide few-shot reasoning—unlike conventional similarity- or diversity-based retrieval.

\paragraph{Cross-lingual transfer (English-only setting).} 
Since English-trained DSRE models (\pare, \cil) lack coverage for Indic scripts, their results are not reported in the table.
Zero-shot prompting yields low scores (typically 20–30 micro-F1), reflecting the limited representation of these languages in the latest LLMs.
Introducing few-shot prompting with English exemplars via \hydra\ improves performance by 8 to 11 average F1 points across LLMs.
Notably, \hydra(Llama-FT) reaches from 24/14 to 35/21 — showing that even English exemplars, when semantically aligned, can substantially aid reasoning in low-resource Indic languages.

% Next, we analyze results for the translate-train setting. For the supervised baseline, we train \parex\ on the translated \textit{X\_train} data for DSRE. For prompting, we use the same LLMs as in Setting 1, and additionally fine-tune Llama3.1-8b on \textit{X\_train}, denoted as $Llama3\text{-}FT_{X}$. We also consider ??? (pending decision on inclusion of CPT).

\paragraph{Translate-train setting.} Training DSRE models on translated data ($X_{\text{train}}$) provides stronger baselines (\parex: 30/20 F1). Zero-shot prompting of open LLMs (Qwen3, Llama3.1) still underperforms these supervised models, but five-shot prompting with \hydra\ yields significant gains (5 to 8 points) over \parex. For \textit{Llama-3-FT} (fine-tuned Llama), the gains are even more pronounced - improving over \parex\ by +17 micro-F1 and over zero-shot Llama-FT by +9 points. Fine-tuned Llama3-FT achieves the best overall performance (47/29 F1), while GPT-4o follows closely (38/34 F1). We also observe interesting language-specific trends (Table \ref{tab:2_lang}): GPT-4o performs best on high-resource scripts like Oriya and Tulu, whereas Llama-FT excels on rarer scripts such as Santhali and Manipuri. A plausible explanation is that Tulu shares its script with Kannada, and GPT-4o has likely encountered both Oriya and Kannada during its pretraining or supervised fine-tuning (SFT) phase. In contrast, Llama fine-tuning remains essential for Santhali and Manipuri, where GPT-4o struggles even with few-shot prompting—likely due to their absence from its pretraining data. 

\paragraph{Translate-test setting.}
When the test data are translated to English, all models exhibit a mild degradation compared to their English performance due to translation noise.
Nevertheless, \hydra\ continues to offer strong improvements: Llama3.1 and Qwen3 gain +15 and +10 micro-F1 points over their zero-shot variants, while GPT-4o achieves 56/54 F1—the highest among all models. These results demonstrate that \hydra\ remains robust across varying LLMs and transfer settings.

\paragraph{\hydra\ vs. other exemplar selection methods:} Across all settings, \hydra\ consistently surpasses other few-shot baselines.
Compared to diversity-based LM-MMR, \hydra(GPT-4o) achieves average gains of +7 micro-F1 on English and +5–6 micro-F1 across Indic languages.
These stem from its hybrid scoring that integrates DSRE confidence with semantic similarity, unlike MMR or TopK-sim, which ignore label information and rely solely on input embedding distance or diversity heuristics. 

\paragraph{Translate-train vs. Translate-test:} Translate-test consistently outperforms translate-train across all LLMs. This is likely attributed to higher translation quality when sentences are translated from target language (X) to English, as opposed to translating English data into the target language.

% We leave a further investigation of the effect of translation noise on the translate-test performance to future work.

\subsection{Ablations}
To understand the contributions of various components to our proposed three-stage algorithm, we perform systematic ablations as follows\footnote{For low-resource languages, the stage 1 and stage 3 ablations are omitted due to excessive prompt length from high token fertility, often exceeding the LLM's maximum sequence length.}:
\begin{itemize}
\itemsep=0em
    \item \textit{w/o candidate relation selection (stage 1)}: Instead of selecting top-$k$ candidate relations in stage 1, we consider all relations in the ontology as candidates. 
    \item \textit{w/o bag selection (stage 2)}: We flatten bags into sentences and assign all bag labels to each of it's sentences. Retrieval is performed directly over sentences.
     \item \textit{w/o sentence selection (stage 3)}: Entire bags (as passages) are provided to the LLM without representative sentence selection.
     % \footnote{For low-resource languages, this ablation is omitted due to excessive prompt length from high token fertility, often exceeding the LLM’s maximum sequence length.}
    \item \textit{w/o semantic similarity}: Retrieval ignores semantic similarity scores; only \pare’s confidence scores are used.
    \item \textit{w/o PARE candidate scores}: \pare\ scores are excluded. Bag selection relies solely on semantic similarity, and sentences within bags are chosen randomly.
    \item \textit{w/o both} (Random): For each candidate label, a random bag containing that label is chosen, followed by a random sentence from it.
     \item \textit{w/o ICL examples}: 
    The LLM selects directly from \pare’s top-5 candidate relations without being shown their exemplars.
\end{itemize}
% \todo{V: check the List of ablations}

\begin{table}[t]
%\vspace*{-0.5ex}
\centering
\small{\begin{tabular}{@{}lrrrrrl@{}}
  \textbf{Ablation} &\textbf{L3.1}   &  \textbf{Qw3} & \textbf{G4o} \\ 
 \hline
 \hydra\  & \textbf{52}/\textbf{47} & \textbf{63}/\textbf{62} & \textbf{63}/60 \\
 w/o sem. sim.  & 45/39 & 60/59 & 60/62 \\
 w/o $f_{PARE}$  & 49/45 & 61/60 & 61/59 \\
 w/o both (Random)  & 45/43 & 56/56 & 60/60 \\
 w/o candidate selection (stage 1) & 39/31 & \textbf{63}/61 & 62/\textbf{64}\\
 w/o bag selection (stage 2) & 48/41 & 62/59 & 56/51  \\
 w/o sentence selection (stage 3) & 46/37 & 60/59 & 59/58 \\
 w/o ICL  & 45/37 & 58/50 & 59/49 \\
 % w/o ontology  & 54/48 & 62/59 \\
\end{tabular}}
\caption{English F1 (micro/macro) scores for different ablation variants of \hydra. Model keys: L3.1 = Llama3.1-8B, Qw3 = Qwen3-235B, G4o = GPT-4o.}
\label{table:ablat1}
%\vspace*{-1ex}
\end{table}
\begin{table}[t]
%\vspace*{-0.5ex}
\centering
\small{\begin{tabular}{@{}lrrrrrl@{}}
  \textbf{Ablation} &\textbf{L3.1} & \textbf{L3.1-FT} & \textbf{Qw3}  & \textbf{G4o} \\ 
 \hline
 % ours & 26 & 43 & 39  \\
 \hydra*  & 32/\textbf{20} & \textbf{45}/\textbf{28} & \textbf{38}/\textbf{31} &  38/34\\
 only sem. sim. & 26/18 & 44/27 & 37/\textbf{31} &  \textbf{39}/34\\
 Random &  26/16 & 38/21 & 37/\textbf{31} & 36/32\\
 w/o bag selection & \textbf{33}/\textbf{20}  & 43/25 & 34/30 & 38/\textbf{35} \\
 (Stage 2) & \\
 w/o ICL & 29/19 & 33/18 & 33/20 & 32/21\\
 % w/o ontology & 24 & 42 &  37\\
\end{tabular}}
\caption{Avg. F1 (micro/macro) over Indic languages for different ablation variants of \hydra\ in \textit{translate-train} setting. *\hydra\ omits semantic retrieval in \textit{translate-train} setting owing to poor performance of retrieval models for low-resource languages (more details in sec. \ref{subsec: sem-x}).}
\label{table:ablat2}
%\vspace*{-1ex}
\end{table}
We present results in tables \ref{table:ablat1} and \ref{table:ablat2} for the monolingual and cross-lingual transfer experiments, respectively. We make the following  observations.

\paragraph{Semantic similarity is most crucial for English, while for low-resource languages \pare\ confidence is critical.} We observe that for English, semantic similarity yields with up to 7 F1 point drop in Llama 3.1, while removing the \pare\ confidence only leads a marginal drop (2 to 3 points) for both LLMs.
In contrast, \parex\ confidence turns to be most critical for cross-lingual setting, with semantic similarity even hurting the performance of \hydra\ (more analysis in \ref{subsec: sem-x} ).
\paragraph{Sentence selection from bags turns out to be crucial step for effective ICL's performance:} When we don't subselect representative sentences from top-k bags, and give entire bags-as-exemplars in prompt to LLM, this hurts the performance by upto 6 micro F1 and 10 macro F1 points as LLMs fail to handle longer context lengths in these cases.
In some sense, \hydra\ is able to exploit a key property of distant supervision -- that all sentences in a bag are not needed for optimal performance. 
\paragraph{W/o Canditate relation selection:}
When we gives exemplars for all 24 relations, it has comparable performance to HYDRE( which uses only top-5 relations) but with almost 5x cost and this method is also not scalable if relation count increases since it requires exemplars for all relations.
% \textbf{Semantic similarity is crucial for GPT-4o but not for the other LLMs:} Without semantic similarity scores for candidate retrieval, GPT-4o's performance hurts by 1 point, while for Llama-3.1 it \red{???}.
\paragraph{ICL exemplars for candidate relations are crucial for both experiments.} We observe that just giving PARE top-5 candidate relations to the LLM is not sufficient as the performance with this approach is lesser than the ICL-based approach by up to 7 points in English and 12 F1 points in translate-train setting. This highlights the significance of carefully selected ICL exemplars in guiding LLM for optimal DSRE performance.

\subsection{Robustness to bag-level evaluation}
In addition to sentence-level evaluation, we seek to evaluate our approach on bag-level test queries as has been done in prior DSRE literature \cite{gao2021manual, rathore-etal-2024-ssp}. This finds applications in scenarios where we have a corpus of documents (e.g., news) and we need to populate a KB.
\begin{figure}
\includegraphics[width=0.45\textwidth, height=6cm]{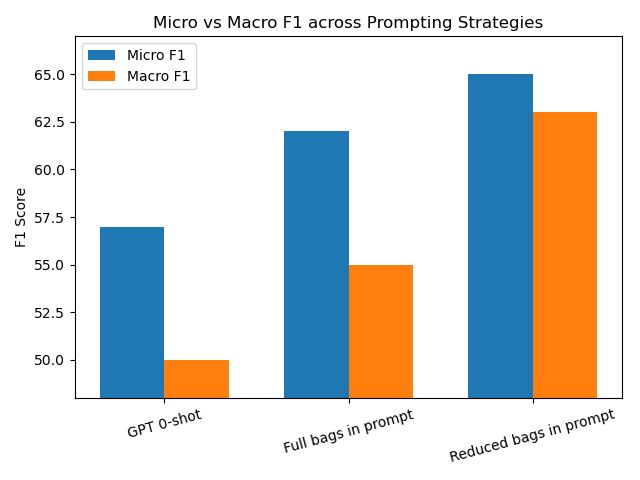}
    \caption{Zero-shot prompting vs full bag ICL vs reduced bag ICL for GPT-4o on English NYT-10m}
    \label{fig:bag_scale}
\end{figure} 

For bag-level queries, we can give bag-level exemplars to LLM by following Algorithm~\ref{alg:selecting_exemplars} till bag selection stage. Then, instead of finding a single representative sentence from each bag, we find a different representative sentence \emph{for each relation} in the bag's labelset. Sentences with highest PARE score for each relation in the bag are kept and rest are discarded. As shown in Fig. \ref{fig:bag_scale}, the \textit{reduced bag} exemplars improve performance by 3 micro F1 and 7 macro F1 over naively giving the full bag exemplars in the prompt. Further this saves cost with around 2$\times$ reduction in prompt tokens.\footnote{Avg. tokens/prompt: full bag - 2451, reduced bag - 1242}
% \subsection{Scalability with few-shot examples}
% We vary the no. of few-shot examples in the range {5, 10, 15, 20}. We observe that ???. 
% \subsection{Error Analysis}
% \subsection{Impact of translation noise on the \textit{translate-test} performance}
% \label{subsec:transtest}
% To assess the impact of translation quality on \textit{translate-test} performance, we examine how noise introduced during translation affects few-shot prompting, and whether our method remains robust under such conditions. Specifically, we translate the query from the target language X to English using the same LLM employed for prompting.

% Given that most LLMs have not encountered X→English translation data for our target languages during supervised fine-tuning (SFT), the resulting translations often suffer from low quality and hallucinations (see examples in ???). We refer to this setup as \textit{translate-test-\textless{}LLM\textgreater{}}.

% We compare the performance of this approach with our main results (Table ???), which use constrained translation via CODEC. The comparative results are presented in Table ???.

% We observe that ???
\subsection{\hydra\ Sensitivity vs $k$}
We analyze \hydra's sensitivity to the number of candidate relations ($k$) selected in Stage 1 before LLM disambiguation. As shown in Figure~\ref{fig:f1_k}, micro-F1 initially improves with increasing $k$, peaking around $k=5$–$10$, after which performance plateaus or slightly declines. In contrast, the underlying DSRE model (PARE) continues to show monotonic gains in Recall$@k$ with increasing $k$ (Figure~\ref{fig:rec_k}), reaching near-perfect recall beyond $k=20$. This discrepancy suggests that presenting too many candidates to LLM may degrade precision likely because LLM receives longer prompts and a complex disambiguation space. So, a moderate $k$ (we set $k=5$ in all our experiments) offers the best trade-off between recall coverage, F1 performance and inference latency.  
% \red{As shown in Table~\ref{table:recall-5}, our DSRE model achieves high Recall@$k$ for values of $k$ significantly smaller than the total number of relation types in the ontology. This not only reduces the number of exemplars needed in the prompt, thereby lowering the cost (prompt length) for querying LLM, but may also improve accuracy by narrowing the focus to likely relevant relations instead of overwhelming the model with all possible ones. We fix $k = 5$ for all our experiments.}
\subsection{Qualitative Analysis}
We conduct confusion analysis of \hydra\ against PARE and 0-shot GPT-4o models for English (see appendix Figure \ref{fig:conf_all}). Further, we manually identify 16 fine-grained relation pairs amongst which models get confused (details in Fig. \ref{fig:relation_matrices_a} and \ref{fig:relation_matrices_b} appendix). We observe that \hydra{} has stronger ability to resolve between subtle close-by relations such as ``Religion" v/s ``Ethnicity" (Ex. \ref{fig:relation_examples_1}, ``Geographic distribution" vs ``Nationality" (Ex. \ref{fig:error_6}), etc.

\subsection{Error Analysis}
We analyze cases where \hydra\ fails to output correct relations (see figures \ref{fig:error_1} to \ref{fig:error_5} appendix). These include (a) errors due to \textit{position bias} (a prevalent bias observed in LLM-as-Judge \cite{zheng2023judging}) towards PARE's top-1 candidate (nearest exemplar label to query) (fig. \ref{fig:error_6}, \ref{fig:error_5}), (b) low recall for multi-label queries (fig. \ref{fig:error_4}), and (c) low recall for specific labels such as ``Ethnicity" (often confused with ``Nationality") (fig. \ref{fig:error_3}), ``Advisors" (vs ``Founders"), etc (detailed analysis in App.  \ref{subsec:error}).

\section{Conclusions and Future Work}
We propose \hydra, a novel hybrid framework that leverages SoTA distantly supervised models for guiding modern day LLMs for the task via efficient In-Context Learning. We show the efficacy of \hydra\ across two experiments: (a) Bag-to-sentence transfer in English, (b) Bag-to-sentence transfer in low-resource languages. We propose effective strategies to adapt \hydra\ to cross-lingual settings under different data availability settings. Ablations show the efficacy of each component in \hydra\ in both monolingual and cross-lingual settings. 
% Further, \hydra\ is also effective in a third experiment, where we provide a bag of sentences for evaluation, instead of a single sentence. This is useful where evidence across multiple sentences can make more reliable predictions.

\hydra\ lays the foundation for further research in DSRE in context of the latest LLMs. Possible future works involve a more complex agentic workflow wherein the agents interact iteratively until convergence to arrive at the correct answer. Moreover, applying \hydra\ to newly added entries in Wikipedia or to local news articles for regional languages would be an interesting and useful future direction.
\section{Limitations}
One potential limitation of our work is the high cost or latency of retrieval from large bags in the training corpus. This problem is escalated further for low-resource languages due to their high token fertility even w.r.t. to latest LLMs. 

Further our framework is not tested for low-resource domains such as biomedical or finance domains involving more complicated semantic relationships between the evolving entities. Similarly, we have only performed experiments on four Indic languages, and have not been able to perform experiments on other low-resource language families due to unavailability of relation extraction data.

\bibliography{custom}
\clearpage
\appendix

\section{Appendix}
\label{sec:appendix}
\subsection{Language details}
Please refer to table \ref{table:lang}. 
\begin{table}[ht]
%\vspace*{-0.5ex}
\centering
\small{\begin{tabular}{@{}lrrrl@{}}
  \textbf{Language} &\textbf{Script} & \textbf{Language code} \\ 
 \hline
 English & Latin & eng\_Latn\\
 Oriya & Oriya & ory\_Orya\\
 Santhali & Ol Chiki & sat\_Olck\\
 Manipuri & Meitei & mni\_Mtei\\
 Tulu & Kannada & tcy\_Tulu\\
\end{tabular}}
\caption{Scripts and language codes for our test languages}
\label{table:lang}
%\vspace*{-1ex}
\end{table}
\subsection{Implementation Details}
We make our code and data public at this link - https://anonymous.4open.science/r/AC2f-pool/. 
\subsubsection{LLMs}
We use (1) unsloth/Meta-Llama-3.1-8B-Instruct-unsloth-bnb-4bit version of Llama 3.1 for local inference as well as for fine-tuning, (2) TogetherAI's unsloth/gemma-3-4b-it-unsloth-bnb-4bit for Gemma3, (3) TogetherAI's Qwen3 235B A22B Instruct 2507 FP8 Throughput for Qwen3 and (4) OpenAI's gpt-4o-2024-05-13 for GPT-4o. 

For local inference and fine-tuning, we use a single NVIDIA A100 40GB GPU node.

\subsubsection{Hyperparameters}
For HYDRE, we set $k$ (no. of exemplars) = 5 following the sensitivity analysis in section \ref{subsec:scale}, model threshold $t$ = 0.5 following \pare's original implementation.  \newline
For LLM inference, we set temperature ($\tau$) = 0.0, max. input length as 2048 tokens and max. output tokens as 256. 

\subsubsection{Baselines}
\label{subsubsec: baselines}
For few-shot prompting baselines, we flatten bags to sentences (assign all bag labels to each of it's sentences) and then perform exemplar selection using (a) Random, (b) Top-k similarity-based, and (c) MMR-based retrieval. \newline
Following \cite{10.1145/3726302.3730194}, we implement MMR using iterative selection as follows: \newline
$ MMR(q, s, S_{t-1}) = \alpha \cdot Sim (q,s) - (1-\alpha) \cdot max_{s' \in S_{t-1}} Sim(s, s') $, \newline
where $S_{t-1}$ denotes the candidate set selected so far at time step $t$.
Here, $\alpha$ trades-off the relevance with diversity and is tuned on English dev set. The optimal value of $\alpha$ is found to be 0.3 and is used across all experiments.
\subsubsection{Fine-tuning details}
\label{subsec:finetune}
\paragraph{Translate-Train: PARE Adaptation:} In the translate-train setting, we first adapt mBERT to each target language $X$ by pretraining it on monolingual corpora derived from \textit{X\_train}. This results in a language-specific variant denoted \texttt{$mBERT_{X}$}. For pretraining, we extend the vocabulary with up to 10,000 randomly initialized new tokens for language X. Subsequently, we fine-tune the PARE model on \textit{X\_train} using $mBERT_X$ as the encoder using a task-specific adapter to obtain \parex.

\vspace{0.5ex}
\noindent
\paragraph{LLaMA 3.1 Fine-Tuning:} We use Low-rank adaptation (LoRA) with $lora\_alpha$ = 64, $lora\_r$ = 16, $lora\_dropout$ = 0.0, Learning rate scheduler as Cosine and warmup ratio as 10\%. For training, we use $per\_device\_train\_batch\_size$ = 8, $gradient\_accumulation\_steps$ = 4 and maximum training steps of 5000 on single NVIDIA A100 40GB GPU. 

The NYT-10m training data consists of 41624 bags and so the number of effective training epochs is = 8*4*5000/41624 $\approx$ 4 epochs.

\vspace{0.5ex}
\noindent
\paragraph{Semantic Retriever Training:} As we do not have an off-the-shelf retriever supporting our target languages, we seek to use a task-specific retriever. Specifically, we leverage the sentence-encoder of \cilx\ for embedding queries and examples with cosine similarity as their similarity scores. 

\subsection{Prompt details}
\label{subsec:prompt}
\textbf{Task Description:} Choose all applicable relations between head and tail entities from the set below. Print each relation in a new line. If none of the relations are applicable, output 'NA'. \\
/people/person/nationality : head entity is a person and tail entity is a country \\
/time/event/locations : head entity is an event and tail entity is a location \\
/people/person/children : head entity is a person and tail entity is another person (child) \\
/business/company/advisors : head entity is a company and tail entity is a person (advisor) \\
/business/location : head entity is a business and tail entity is a location \\
/business/company/majorshareholders : head entity is a company and tail entity is a person or organization (major shareholder) \\
/people/person/place\_lived : head entity is a person and tail entity is a location \\
/business/company/place\_founded : head entity is a company and tail entity is a location \\
/location/neighborhood/neighborhood\_of : head entity is a neighborhood and tail entity is a larger location (city, town) \\
/people/deceasedperson/place\_of\_death : head entity is a deceased person and tail entity is a location \\
/film/film/featured\_film\_locations : head entity is a film and tail entity is a location \\
/location/region/capital: head entity is a region and tail entity is a city (capital) \\
/business/company/founders : head entity is a company and tail entity is a person (founder) \\
/people/ethnicity/geographic\_distribution : head entity is an ethnicity and tail entity is a location where the ethnicity is commonly found \\
/location/country/administrative\_divisions : head entity is a country and tail entity is a subdivision (state, province) \\
/people/deceasedperson/place\_of\_burial : head entity is a deceased person and tail entity is a burial site \\
/location/country/capital : head entity is a country and tail entity is a city (capital) \\
/business/person/company : head entity is a person and tail entity is a company they are associated with \\
/location/location/contains : head entity is a larger location and tail entity is a smaller location within it \\
/location/administrative\_division/country : head entity is an administrative division (state, province) and tail entity is a country \\
/location/us\_county/county\_seat: head entity is a U.S. county and tail entity is the county seat \\
/people/person/religion: head entity is a person and tail entity is a religion \\
/people/person/place\_of\_birth : head entity is a person and tail entity is a location (birthplace) \\
/people/person/ethnicity: head entity is a person and tail entity is an ethnicity \\
NA : no relation from the set exists between the given entity pair
\newline
\textbf{Input format:} \newline
Input: \{sentence\}\newline
\textbf{Output format:} \newline
Output: \{relation\} \newline
\textbf{Verbalizer: } \newline
Extract the relation based on exact string match \newline
A sample format of input and output is shown in Figure \ref{fig:relation_examples_1}. 

\subsection{HYDRE Algorithm}
Please refer to Algorithm \ref{alg:selecting_exemplars}.
\begin{algorithm*}[ht]
\small
\caption{Exemplar Selection for HYDRE}
\label{alg:selecting_exemplars}
\textbf{Input:} Query sentence $q$; bags of sentences $\mathcal{B} = \{B_j\}$; trained DSRE model scores $f(s, r)$ for sentence $s$ and relation $r$; semantic similarity function $\mathrm{sim}(q, B_j)$; confidence threshold $t$; number of exemplars $k$.\\
\textbf{Output:} Ordered list of selected exemplar sentences.

\begin{algorithmic}[1]
\State Compute $f(q, r)$ for all relations $r$ in the ontology.
\State $\mathcal{R}' \gets$ top-$k$ relations ranked by $f(q, r)$ \Comment{Candidate relation selection}

\State Initialize exemplar list $\mathcal{E} \gets [\,]$.
\For{each $r \in \mathcal{R}'$}
    \State $\mathcal{B}_r \gets \{B_j \in \mathcal{B} \mid r \in \textit{labelset}(B_j)\}$.
    \State $B_r \gets \arg\max_{B_j \in \mathcal{B}_r} [\mathrm{sim}(q, B_j) + f_{PARE}(B_j, r)]$.
        \Comment{Select most relevant bag for $r$}
    \State For each $s \in B_r$, compute $c(s) = \sum_{r_a \in \text{labels}(B_r)} \mathbb{I}[f(s, r_a) > t]$.
    \State Let $v_{\max} \gets \max_{s \in B_r} c(s)$. \Comment{max. no. of relations with confidence > threshold}
    \State $\mathcal{S}' \gets \{s \in B_r \mid c(s) = v_{\max}\}$. \Comment{candidate sentences with max. label coverage}
    \State $s^* \gets \arg\max_{s \in \mathcal{S}'} \sum_{r_a \in \text{labels}(B_r)} f(s, r_a)$.
        \Comment{Pick sentence with highest aggregate confidence}
    % \State Add $(s^*, r_a, f(q, r))$ to $\mathcal{E}$.
    \State Add $(s^*, r_a)$ to $\mathcal{E}$.
\EndFor
\State Sort $\mathcal{E}$ in ascending order of $f(q, r)$.
    \Comment{keep most informative examples at the last (closer to query)}
\State \Return Ordered list of exemplar sentences $\{s^*\}$ from $\mathcal{E}$.
\end{algorithmic}
\end{algorithm*}

\subsection{Detailed Language-wise results}
Please refer to tables \ref{tab:1_lang}, \ref{tab:2_lang} and \ref{tab:3_lang}.
\begin{table*}[ht]  % Use [t!] for better placement control
\centering
\small
\begin{tabular}{@{}lrrrrrrr@{}}
\toprule
\textbf{Model} & \textbf{eng\_Latn} & \textbf{ory\_Orya} & \textbf{sat\_Olck} & \textbf{mni\_Mtei} & \textbf{tcy\_Tulu} & \textbf{Avg. (Micro)\textsuperscript{\textdagger}} & \textbf{Avg. (Macro)\textsuperscript{\textdagger}} \\ 
\midrule
\textit{Supervised} \\
% HICLRE \cite{li2022hiclre} & 31/18 & --  & -- & -- & -- & -- & -- \\
% HFMRE \cite{li2023hfmre} & 32/18 & --  & -- & -- & -- & -- & -- \\
\pare\ \cite{rathore2022pare} & 42/31 & --  & -- & -- & -- & -- & -- \\
\cil\ \cite{chen-etal-2021-cil} & 43/32 & -- & -- & -- & -- & -- & -- \\
\hline
\textit{zero-shot} \\
%Gemma3\text{-}4b &19/17 & 15/13 & 15/14 & 10/10 & 18/15 & 15 & 13 \\
Qwen3-235B-A22B & 49/40 & 46/40 & 6/5 & 2/1 & 45/39 & 25 & 21\\
Llama3.1\text{-}8b & 31/17 & 26/21 & 21/14 & 10/7 & 29/21 & 22 & 16\\
Llama3.1-8B-FT$_{En}$ & 60/44 & 33/23 & 15/8 & 11/4 & 36/21 & 24 & 14\\
GPT-4o & 56/57 &  56/56 & 10/5 & 11/7 & 55/\textbf{55} & 33 & 31\\
\hline
\hline
\midrule
\textit{5-shot} \\
%HYDRE(Gemma3-4B) & 24/19 & 21/19 & 19/17 & 4/2 & 21/18 & 16 & 14 \\
HYDRE(Qwen3-235B-A22B) & \textbf{63*}/\textbf{62} & 52/46 & 9/7 & 2/2 & 53/48 & 29 & 26 \\
HYDRE(Llama3.1-8B) & 52/47 & 37/24 & \textbf{24*}/\textbf{16} & \textbf{21*}/\textbf{11} & 38/25 & 30 & 19\\
HYDRE(Llama3.1-8B-FT$_{En}$) & 61/45 & 47/32 & \textbf{24*}/13 & \textbf{21*}/8 & 49/31 & 35 & 21\\
HYDRE(GPT-4o) & \textbf{63*}/60 & \textbf{57}/\textbf{57} & 18/10 & 11/8 & \textbf{56}/\textbf{55} & \textbf{36*} & \textbf{33}\\
\bottomrule
\end{tabular}
\caption{Results for English-only data setting. In each entry, we report micro and macro F1 scores. \textdagger The reported average scores are over non-English languages. * McNemar's p-value $<10^{-5}$ (valid for micro-F1 comparison).}
\label{tab:1_lang}
\end{table*}

\begin{table*}[ht]  % Use [t!] for better placement control
\centering
\small
\begin{tabular}{@{}lrrrrrrr@{}}
\toprule
\textbf{Model} & \textbf{ory\_Orya} & \textbf{sat\_Olck} & \textbf{mni\_Mtei} & \textbf{tcy\_Tulu} & \textbf{Avg. (Micro)} & \textbf{Avg. (Macro)} \\ 
\midrule
\textit{Supervised} \\
\parex\ & 31/20 & 29/20 & 28/19 & 30/19 & 30 & 20 \\
\cilx\ & 29/18 & 22/15 & 25/19 & 29/20 & 26 & 18\\
\hline
\textit{zero-shot} \\
%Gemma3-4b  & 15/13 & 15/14 & 10/10 & 18/15 & 15 & 13\\
Qwen3-235B-A22B & 46/40 & 6/5 & 2/1 & 45/39 & 25 & 21\\ 
Llama3.1-8b  & 26/21 & 21/14 & 10/7 & 29/21 & 22 & 16 \\
GPT-4o & 56/\textbf{56} & 8/6 & 8/6 & 56/53 & 32 & 30\\
Llama3.1-8B-FT$_{En}$ & 33/23 & 10/7 & 2/1   & 36/21 & 20 & 13 \\ 
Llama3.1-8B-FT$_{X}$ & 51/37 & 33/18 & 24/11 & 44/26 & 38  & 23 \\
%\llamacptftx & -- & -- & -- & -- & -- \\
\hline
\hline
\midrule
\textit{5-shot} \\
%HYDRE$_{X}$(Gemma3-4B)  & 21/16 & 22/17 & 8/5  & 25/18 & 19 & 14 \\
HYDRE$_{X}$(Qwen3-235B-A22B) & 56/53 & 23/12 & 14/6 & 57/52 & 38 & 31\\
HYDRE$_{X}$(Llama3.1-8B)  & 38/22 & 34/19 & 25/13 & 42/28 & 35 & 21 \\
HYDRE$_{X}$(GPT-4o) & \textbf{58*}/\textbf{56} & 17/11 & 20/15 & \textbf{57}/\textbf{55} & 38 & \textbf{34} \\
HYDRE$_{X}$(Llama3.1-8B-FT$_{En}$) & 48/33    & 27/15 & 21/8  & 47/31 & 36 & 22 \\
HYDRE$_{X}$(Llama3.1-8B-FT$_{X}$) & 56/39    & \textbf{45*}/\textbf{24}   & \textbf{35*}/\textbf{16}   & 51/36 & \textbf{47*} & 29\\
%\llamacptparetopkx & -- & -- & -- & -- & -- \\
\bottomrule
\end{tabular}
\caption{Results for Translate-train setting. * McNemar's p-value $<10^{-5}$.}
\label{tab:2_lang}
\end{table*}

\begin{table*}[ht]  % Use [t!] for better placement control
\centering
\small
\begin{tabular}{@{}lrrrrrr@{}}
\toprule
\textbf{Model} & \textbf{ory\_Orya} & \textbf{sat\_Olck} & \textbf{mni\_Mtei}  & \textbf{tcy\_Tulu} & \textbf{Avg. (Micro)} & \textbf{Avg. (Macro)} \\ 
\midrule
\textit{Supervised} \\
\pare\ & 35/25 & 31/21 & 33/25 & 32/22 & 33 & 23\\
\cil\ & 34/23 & 33/21 & 34/25 & 36/25 & 34 & 24 \\
\hline
\textit{zero-shot} \\
%Gemma3\text{-}4b & 18/15 & 19/16 & 19/16 & 21/17 & 19 & 16 \\
Qwen3-235B-A22B & 44/37 & 43/34 & 43/35 & 44/34 & 44 & 35\\
Llama3.1  & 29/25 & 26/22 & 29/25 &  30/26 & 29 & 25 \\
Llama3.1-8B-FT$_{En}$ & 49/35 & 45/29 & 49/35 & 52/39 & 49 & 34 \\
GPT-4o & 53/51& 50/45 &55/\textbf{57} & 56/51 & 54 & 51\\
\hline
\hline
\midrule
\textit{5-shot} \\
%HYDRE(Gemma3-4B)  & 20/15 & 18/13 & 20/19 & 21/15 & 20 & 16\\
HYDRE(Qwen3-235B-A22B) & 52/46 & 51/42 & 55/47 & 56/48 & 54 & 46\\
HYDRE(Llama3.1-8B)  & 46/38 & 41/31 & 46/38 & 44/38 & 44 & 36 \\
HYDRE(Llama3.1-8B-FT$_{En}$) & 51/37 & 50/33 & 50/37 & 54/40 & 51 & 37 \\
HYDRE(GPT-4o) & \textbf{55*}/\textbf{54} & \textbf{53*}/\textbf{49} & \textbf{56}/\textbf{57}& \textbf{59*}/\textbf{55} & \textbf{56*} & \textbf{54}\\
\bottomrule
\end{tabular}
\caption{Results for Translate-test setting. * McNemar's p-value $<10^{-5}$.}
\label{tab:3_lang}
\end{table*}

\subsection{Confusion Analysis}
\label{subsec:conf}
\begin{figure}[ht]
\includegraphics[width=0.95\columnwidth, height=8cm]{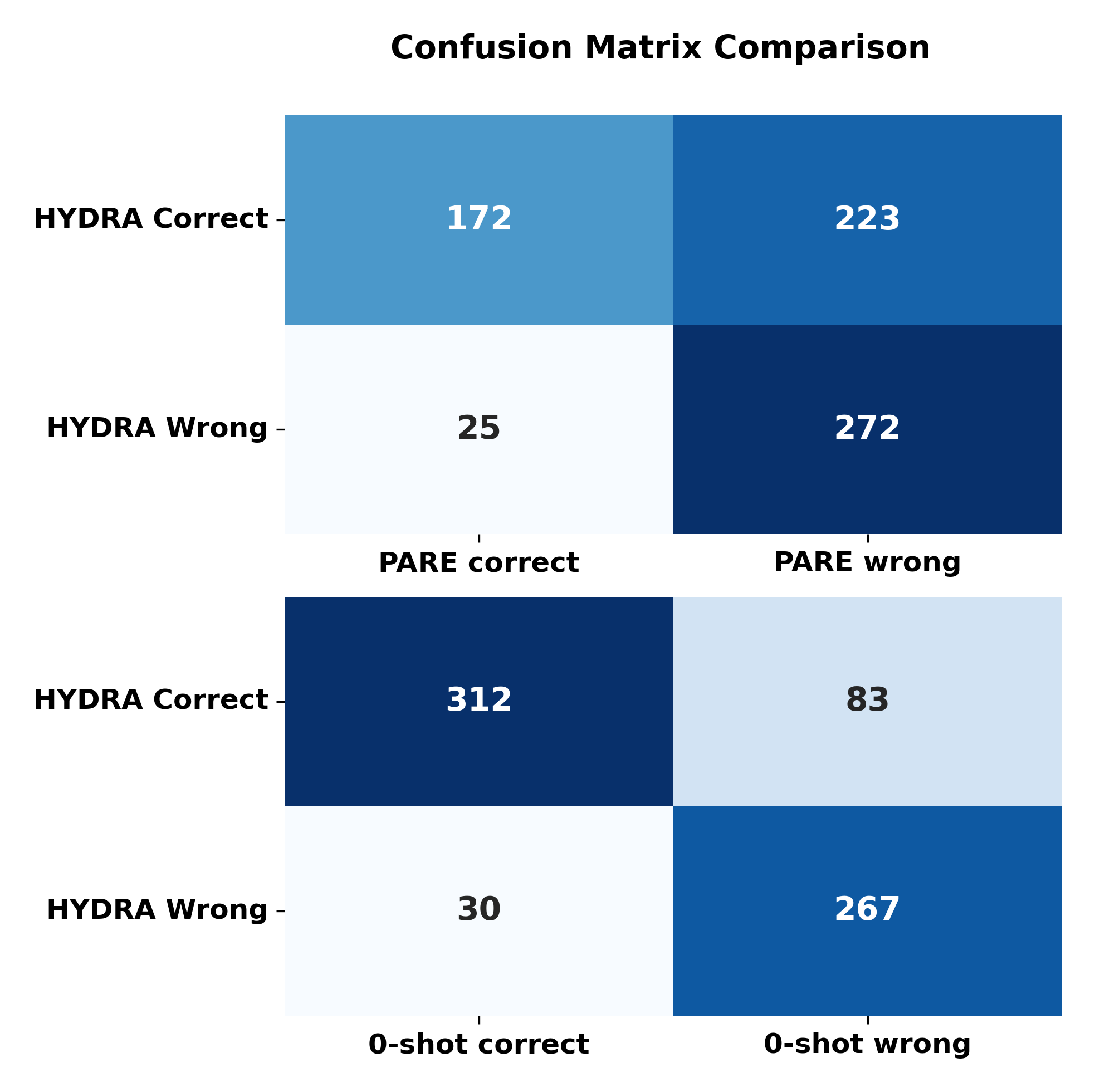}
    \caption{Confusion analysis of \hydra\ vs the baselines (aggregated over all relations)}
    \label{fig:conf_all}
\end{figure}

\begin{figure*}[t]
    \centering
    \captionsetup[subfigure]{labelformat=simple, labelsep=space, font=small}
    
    % First row
    \begin{subfigure}[t]{0.48\linewidth}
        \includegraphics[width=\linewidth, valign=t]{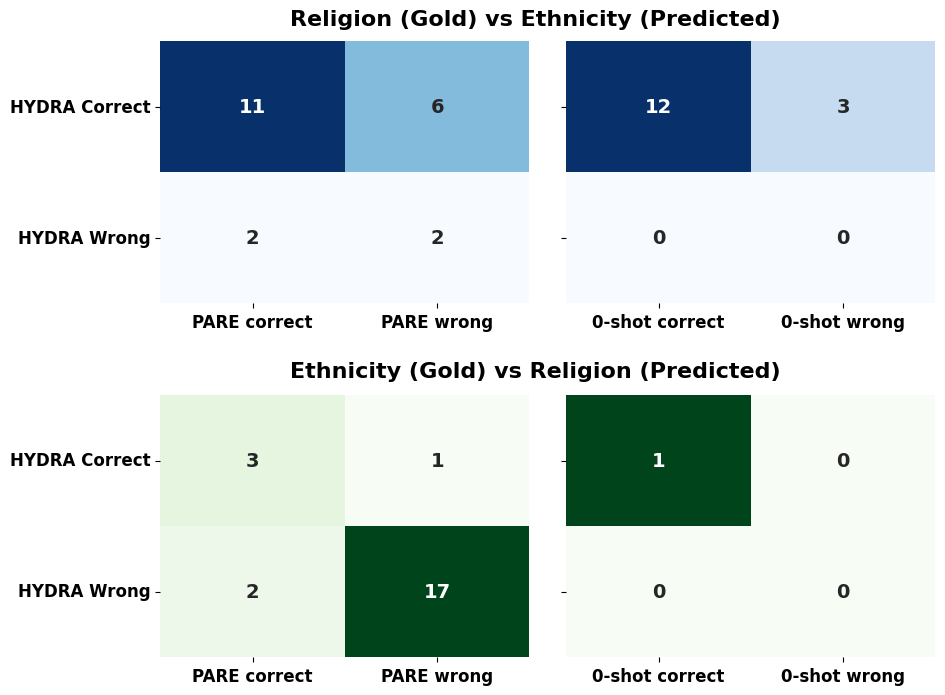}
        \caption{Religion vs Ethnicity}
        \label{fig:conf_a}
    \end{subfigure}
    \hfill
    \begin{subfigure}[t]{0.48\linewidth}
        \includegraphics[width=\linewidth, valign=t]{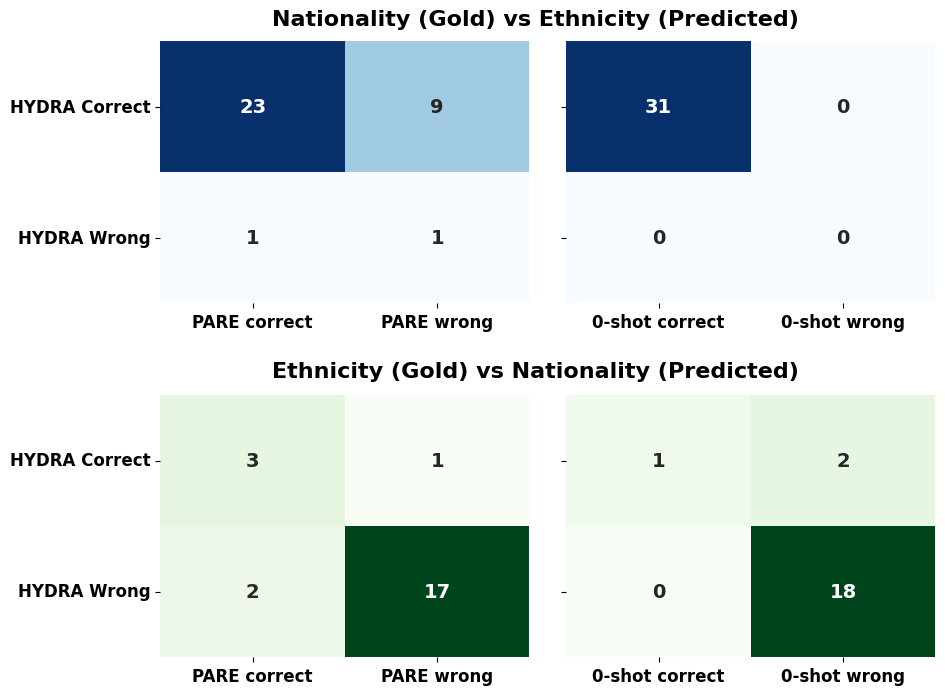}
        \caption{Nationality vs Ethnicity}
        \label{fig:conf_b}
    \end{subfigure}
    
    \vspace{0.3cm} % Reduced vertical spacing
    
    % Second row
    \begin{subfigure}[t]{0.48\linewidth}
        \includegraphics[width=\linewidth, valign=t]{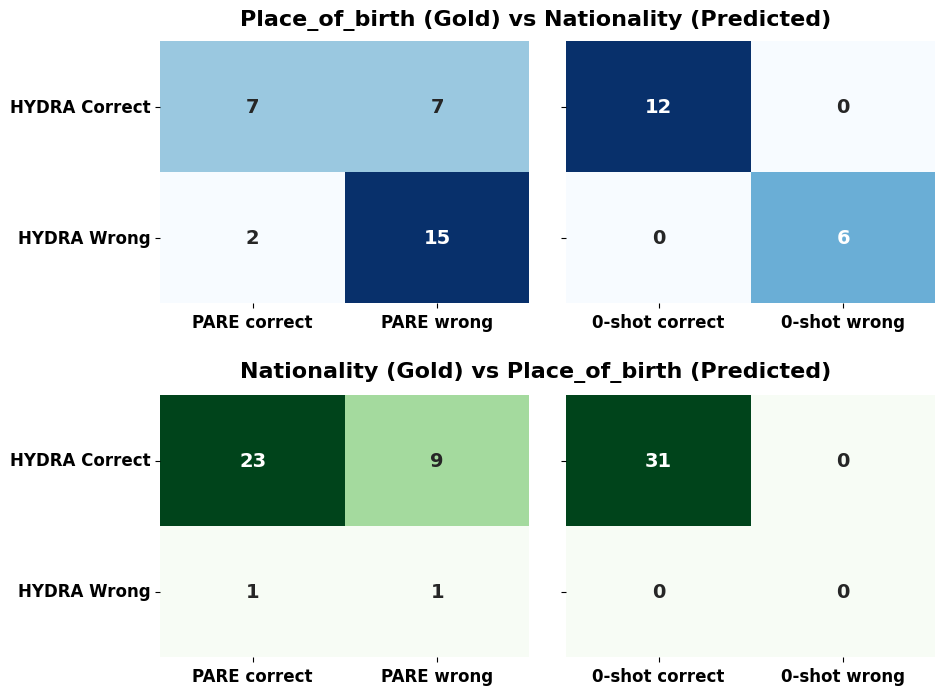}
        \caption{Nationality vs Place of Birth}
        \label{fig:conf_c}
    \end{subfigure}
    \hfill
    \begin{subfigure}[t]{0.48\linewidth}
        \includegraphics[width=\linewidth, valign=t]{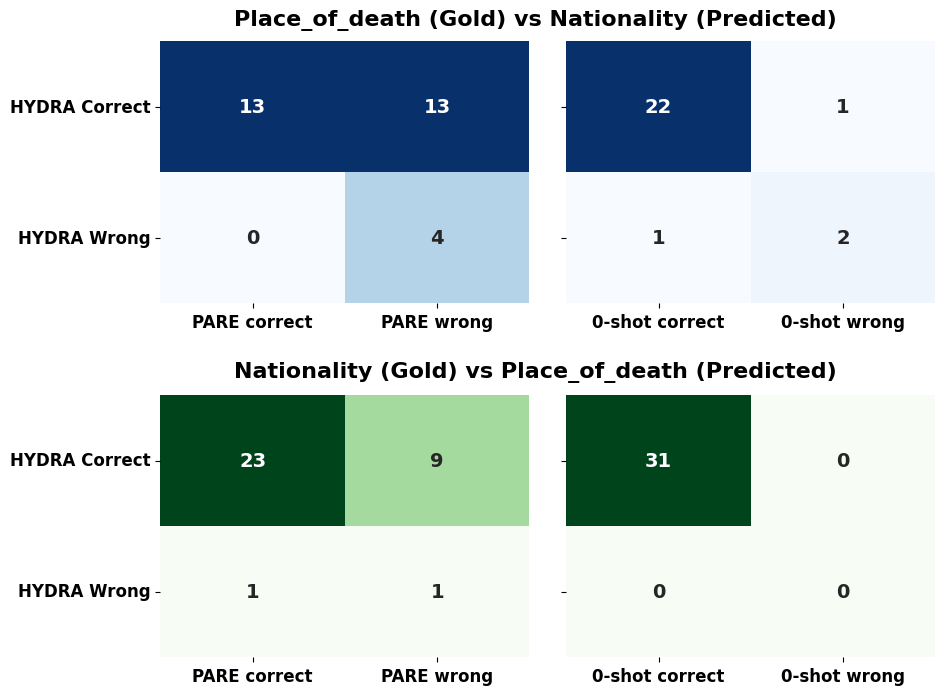}
        \caption{Nationality vs Place of Death}
        \label{fig:conf_d}
    \end{subfigure}

    % 3rd row
    \begin{subfigure}[t]{0.48\linewidth}
        \includegraphics[width=\linewidth, valign=t]{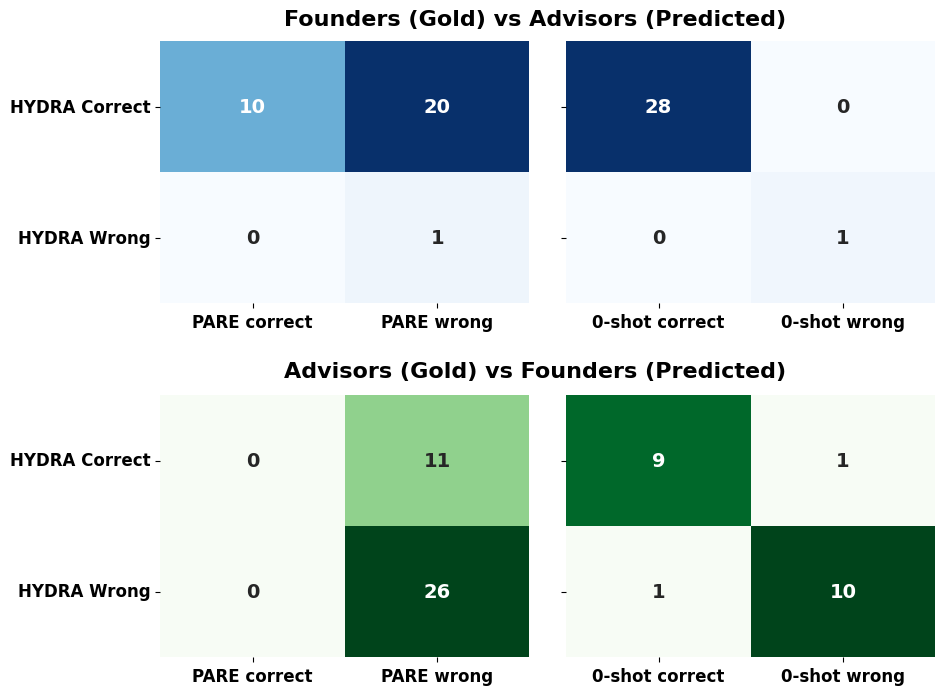}
        \caption{Company founders vs Company Advisors}
        \label{fig:conf_e}
    \end{subfigure}
    \hfill
    \begin{subfigure}[t]{0.48\linewidth}
        \includegraphics[width=\linewidth, valign=t]{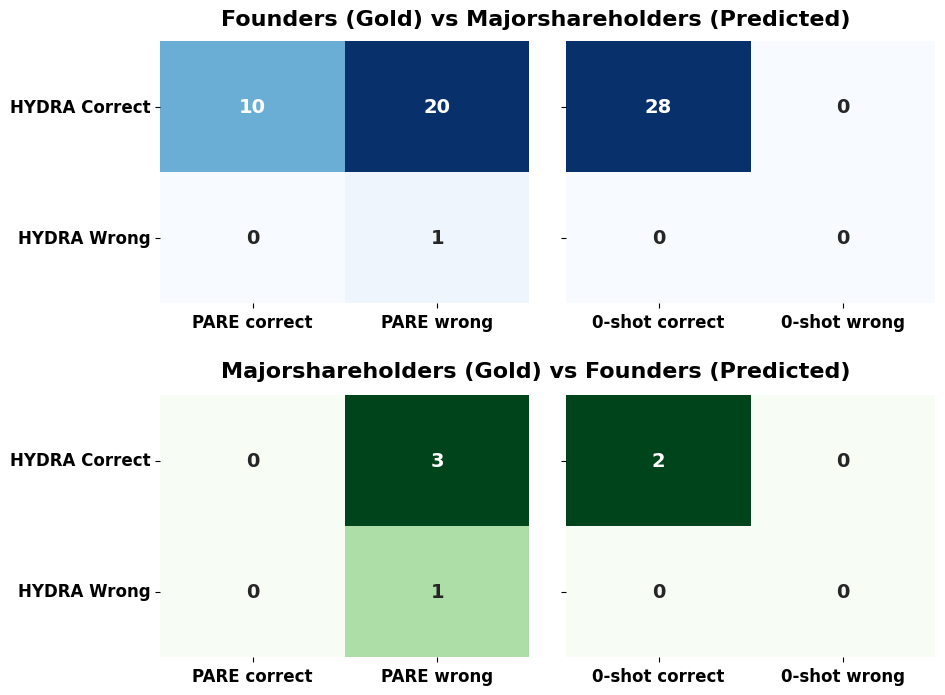}
        \caption{Company founders vs Company Majorshareholders}
        \label{fig:conf_f}
    \end{subfigure}
    \caption{Confusion matrices analyzing \hydra\ vs the baselines across relation types}
    \label{fig:relation_matrices_a}
\end{figure*}

\begin{figure*}[t]
    \centering
    \captionsetup[subfigure]{labelformat=simple, labelsep=space, font=small}
     % 4th row
    \begin{subfigure}[t]{0.48\linewidth}
        \includegraphics[width=\linewidth, valign=t]{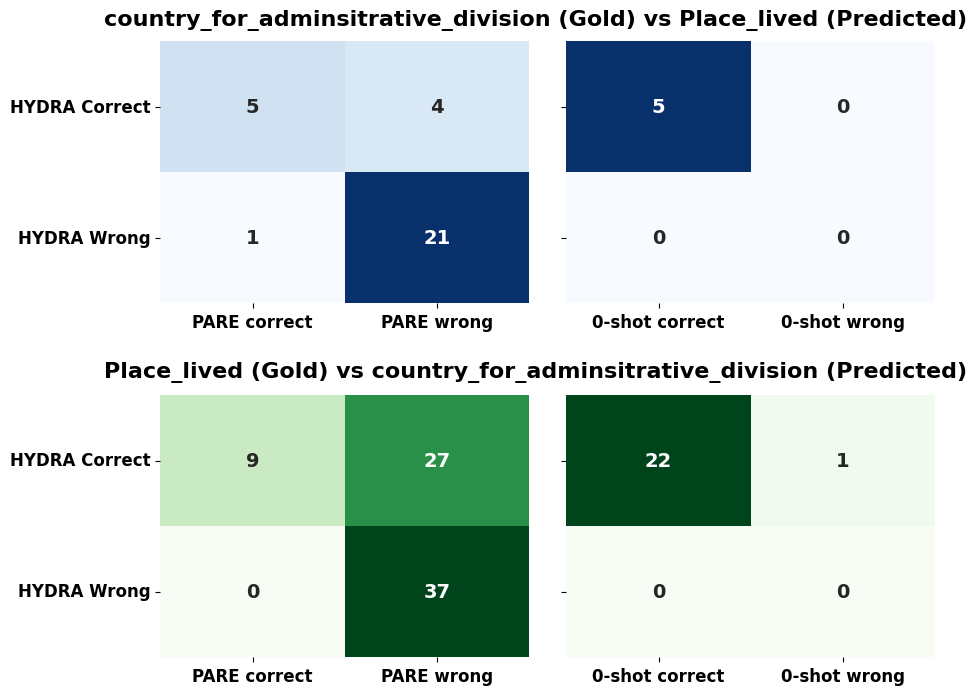}
        \caption{Country for Administrative divisions vs Place lived}
        \label{fig:conf_g}
    \end{subfigure}
    \hfill
    \begin{subfigure}[t]{0.48\linewidth}
        \includegraphics[width=\linewidth, valign=t]{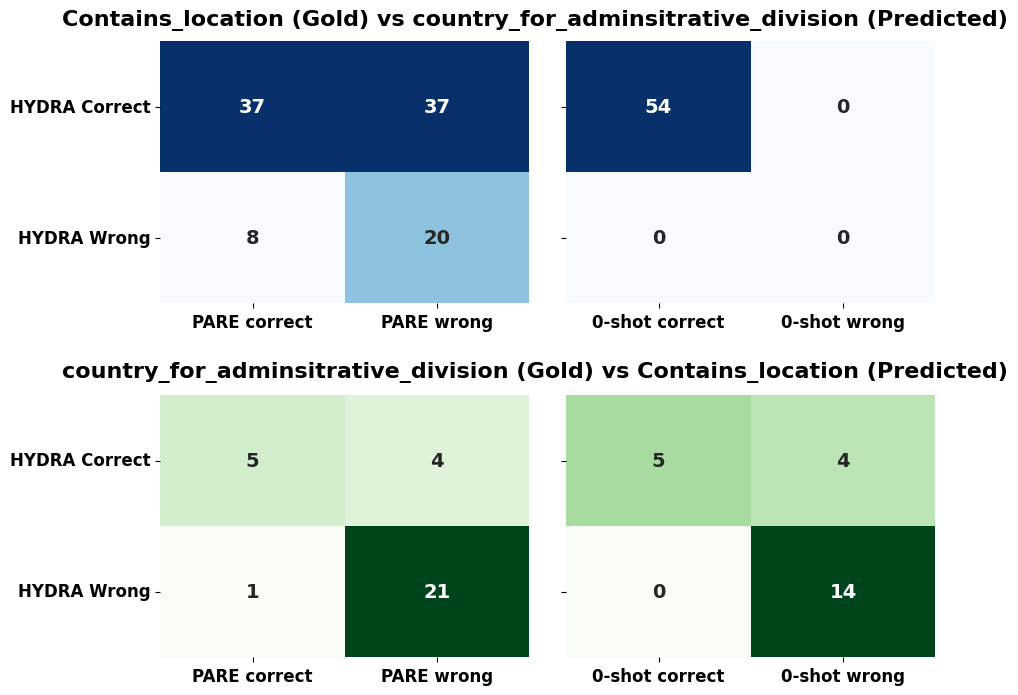}
        \caption{Country for Administrative divisions vs Contains}
        \label{fig:conf_h}
    \end{subfigure}
    % 5th row
    \begin{subfigure}[t]{0.48\linewidth}
        \includegraphics[width=\linewidth, valign=t]{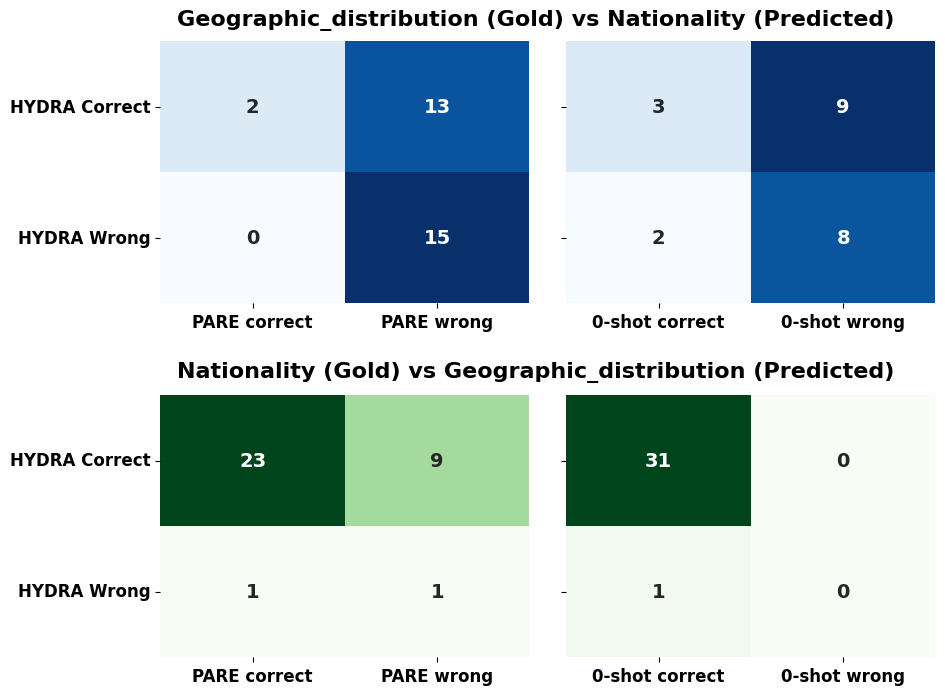}
        \caption{Nationality vs Geographic distribution}
        \label{fig:conf_i}
    \end{subfigure}
    \hfill
    \begin{subfigure}[t]{0.48\linewidth}
        \includegraphics[width=\linewidth, valign=t]{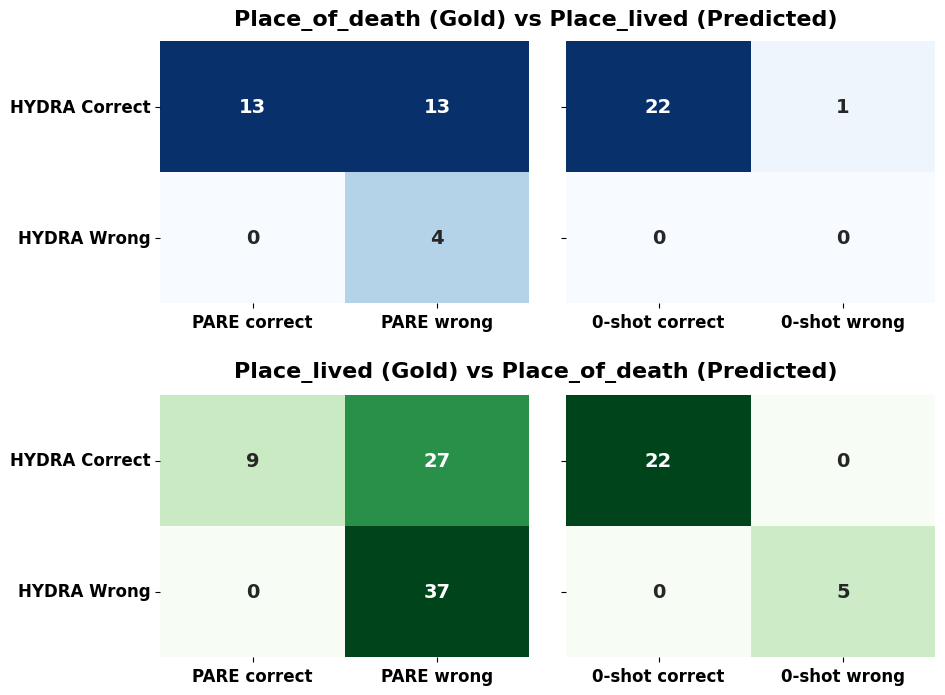}
        \caption{Place lived vs Place of Death}
        \label{fig:conf_j}
    \end{subfigure}
     % 6th row
    \begin{subfigure}[t]{0.48\linewidth}
        \includegraphics[width=\linewidth, valign=t]{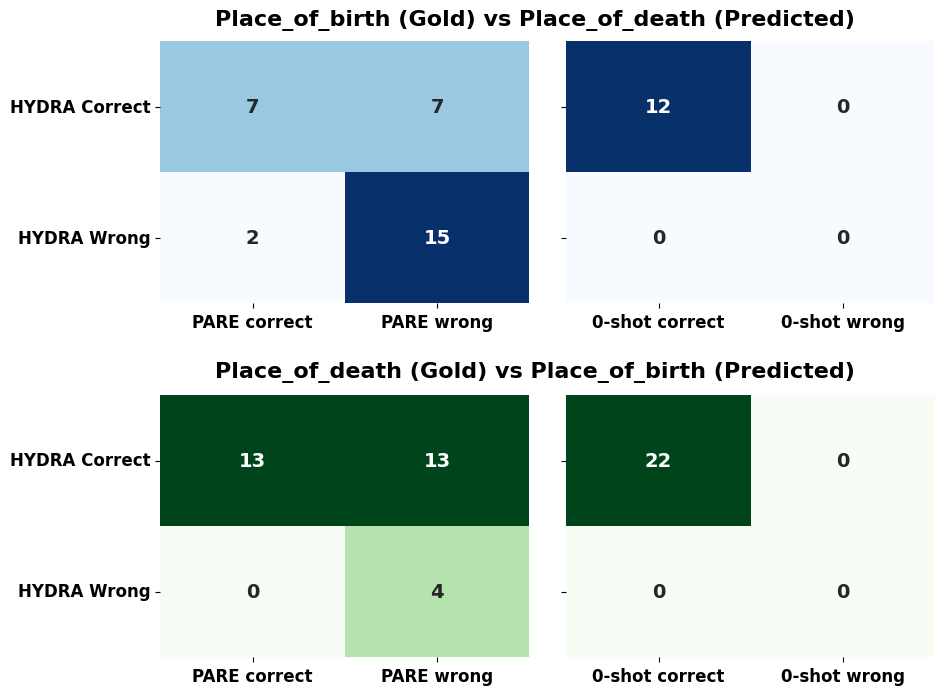}
        \caption{Place of Birth vs Place of Death}
        \label{fig:conf_k}
    \end{subfigure}
    \hfill
    \begin{subfigure}[t]{0.48\linewidth}
        \includegraphics[width=\linewidth, valign=t]{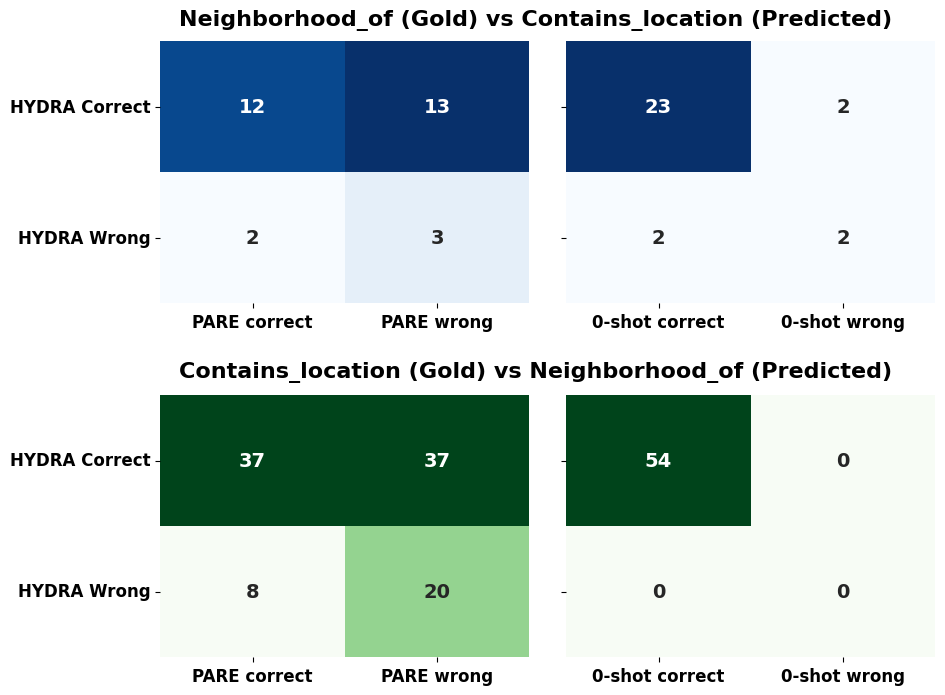}
        \caption{Neighborhood\_of vs Contains\_location}
        \label{fig:conf_l}
    \end{subfigure}
    \caption{Confusion matrices analyzing \hydra\ vs the baselines across relation types}
    \label{fig:relation_matrices_b}
\end{figure*}

We present confusion matrix (aggregated over all labels) in Figure \ref{fig:conf_all}. We manually identify 16 categories of relation pairs amongst which confusion is observed and present their confusion matrices in Figures \ref{fig:relation_matrices_a} and \ref{fig:relation_matrices_b}. We depict some qualitative examples, in which \hydra\ correctly identifies the relation while 0-shot and PARE make a mistake, in the following subsection.
\subsubsection{Qualitative examples}
\label{subsubsec:qual}
We select examples which are correctly predicted by \hydra\ but misclassified by 0-shot prompting in order to study the impact of ICL examples. An example is shown in figure \ref{fig:relation_examples_1} where \hydra\ correctly predicts "Religion" as opposed to "Ethnicity" by 0-shot prompting. ICL examples depict that "Ethnicity" is PARE's top-ranked candidate (closest to the query) while "Religion" is ranked second and \hydra\ is able to subtly distinguish between the 2 based on their exemplars. Example \ref{fig:relation_examples_2} shows another case where \hydra\ correctly predicts "Advisors" v/s "Company" predicted by 0-shot. This is a case where the predicted label "Advisors" is not listed in PARE's top-5 candidates and despite this the \hydra\ is able to predict it. This is attributed to ICL's ability to predict labels beyond what are included in the in-context examples. Other scenarios where \hydra\ dominates 0-shot prompting include multi-label prediction as depicted in Ex. \ref{fig:relation_examples_4}, where \hydra\ predicts "/place\_of\_burial" as well as "/place\_lived" while 0-shot misses the obvious label "/place\_lived" while only predicting "/place\_of\_burial". But this is not always the case as shown in Ex. \ref{fig:error_3}, where PARE top-1 is correct label (" /administrative\_division/country") but \hydra\ predicts a wrong one (" /location/location/contains") which is not in PARE top-5 (ICL Examples). Ex. \ref{fig:error_4} depicts multi-label query where \hydra\ predicts 2 out of 3 labels correctly while 0-shot predicts only one of those. Ex. \ref{fig:error_5} is another case where \hydra\ correctly predicts a more nuanced label "/place\_founded" but misses an obvious label "/business/location", which is not covered in ICL exemplars.

\subsection{Error Analysis}
\label{subsec:error}
We next analyze cases where \hydra\ fails to recall correct relations. One case is where \hydra\ gets biased to simply predict the PARE's top-1 candidate and misses others in multi-label queries. An example is shown in Figure \ref{fig:error_1} where \hydra\ predicts "/geographic\_dsitribution" (PARE's top-1) while missing the label "/nationality" (PARE's top-3), which is predicted by 0-shot prompting.\newline
We further divide the testset facts into 2 categories - (a) for which gold label is in PARE's top-5, and (b) gold label is out of PARE top-5. The individual micro F1 scores on both categories is presented in Table \ref{table:error}. We observe that missing the gold label in PARE top-5 significantly hurts the downstream LLM F1 score compared to hitting it. This shows the positive impact of candidate selection step on the downstream performance of \hydra..
\begin{table}[ht]
%\vspace*{-0.5ex}
\centering
\small{\begin{tabular}{@{}lrrrl@{}}
  \textbf{Category} &\textbf{No. of Facts} & \textbf{micro F1 score} \\ 
 \hline
 gold label in top-5    & 594 & 67\\
 gold label out of top-5    & 98 & 34\\
\end{tabular}}
\caption{micro F1 scores for test-set categories}
\label{table:error}
%\vspace*{-1ex}
\end{table}

\begin{figure}[ht]
\centering
% Example Box
\begin{examplebox}
\textbf{Input}: But Mr. Wallace , 45 , ... in partnership for a time with \texttt{<Tail>} Mark Thatcher \texttt{</Tail>} , son of the former British Prime Minister \texttt{<Head>} Margaret Thatcher \texttt{</Head>} , has detractors , ... . \\
\textbf{Output}: /children \\

\textbf{Input}: In the wake of the torture and killing in February of \texttt{<Head>} Ilan Halimi \texttt{</Head>} , a 23-year-old Jew , attention has focused on an undeniable problem : anti-Semitism among \texttt{<Tail>} France \texttt{</Tail>} 's second-generation immigrant youth , ... . \\
\textbf{Output}: /nationality \\

\textbf{Input}: Another concern is that \texttt{<Tail>} Ethiopia \texttt{</Tail>} and Eritrea , bitter enemies that recently fought ... , with Eritrea suspected of ... to support the Islamists and \texttt{<Head>} Ethiopian \texttt{</Head>} officials now admitting , ... . \\
\textbf{Output}: /geographic\_distribution \\

\textbf{Input}: The case has fueled feelings here of an assault against \texttt{<Tail>} Islam \texttt{</Tail>} , coming after ... and , more recently , cartoons ... that mocked the Prophet \texttt{<Head>} Muhammad \texttt{</Head>} . \\
\textbf{Output}: /religion \\

\textbf{Input}: ... , Mr. Delli Colli worked with generations of \texttt{<Tail>} Italian \texttt{</Tail>} directors , including \texttt{<Head>} Pier Paolo Pasolini \texttt{</Head>} , ... . \\
\textbf{Output}: /ethnicity
\end{examplebox}

% Query Box
\begin{querybox}
\textbf{Input}: A jury convicted a 27-year-old British \texttt{<Tail>} Muslim \texttt{</Tail>} , \texttt{<Head>} Umran Javed \texttt{</Head>} , of soliciting murder and inciting racial hatred ... .
\end{querybox}

% Output Box
\begin{outputbox}
\hydra: \correct{/religion} \\
GPT-4o 0-shot: \wrong{/ethnicity} 
\end{outputbox}
\caption{\textbf{Example 4}: \hydra\ correctly predicts the \textit{``Religion"} label while 0-shot confuses it with \textit{``Ethnicity''}.}
\label{fig:relation_examples_1}
\end{figure}
\begin{figure}[ht]
\centering

% Example Box
\begin{examplebox}
\textbf{Input}: Mr. White , 54 , of \texttt{<Tail>} Centerport \texttt{</Tail>} , has held top environmental posts with \texttt{<Head>} Suffolk County \texttt{</Head>} and the Town of Huntington . \\
\textbf{Output}: /location/location/contains \\

\textbf{Input}: Being able to combine Loudeye services with \texttt{<Head>} Nokia \texttt{</Head>} terminals provides ... , '' said Ilkka Raiskinen , senior vice president for multimedia experiences at Nokia , based in \texttt{<Tail>} Espoo \texttt{</Tail>} , Finland . \\
\textbf{Output}: /place\_founded \\

\textbf{Input}: When people say `...', Jerry was just as responsible for that as my dad , '' ' said \texttt{<Tail>} Brian Henson \texttt{</Tail>} , \texttt{<Head>} Jim Henson \texttt{</Head>} 's son , who serves with his sister Lisa as chairman and chief executive of the Jim Henson Company . \\
\textbf{Output}: /children \\

\textbf{Input}: ... , among them Larry Page and \texttt{<Tail>} Sergey Brin \texttt{</Tail>} , the co-founders of \texttt{<Head>} Google \texttt{</Head>} , who bought ... . \\
\textbf{Output}: /business/company/founders, 
/business/company/majorshareholders \\

\textbf{Input}: The service has more ways to shield users ' identities .... , said \texttt{<Tail>} Antony Brydon \texttt{</Tail>} , \texttt{<Head>} Visible Path \texttt{</Head>} 's chief executive , above . '' \\
\textbf{Output}: /business/company/founders
\end{examplebox}

% Query Box
\begin{querybox}
\textbf{Input}: `` \texttt{<Head>} MySpace \texttt{</Head>} is dedicated to ensuring that ..., '' \texttt{<Tail>} Chris DeWolfe \texttt{</Tail>} , the chief executive of MySpace , said in a statement .
\end{querybox}

% Output Box
\begin{outputbox}
\hydra: \correct{/business/company/advisors} \\
0-shot: \wrong{/business/person/company}
\end{outputbox}
\caption{\textbf{Example 5}: \hydra\ correctly predicts the \textit{``Advisors"} label while 0-shot misses it.}
\label{fig:relation_examples_2}
\end{figure}
\begin{figure}[ht]
\centering
% Example Box
\begin{examplebox}
\textbf{Input}: Mr. Marek started the \texttt{<Head>} Fairfield \texttt{</Head>} Theater Company in 2001 ... at \texttt{<Tail>} Fairfield University \texttt{</Tail>} . \\
\textbf{Output}: /location/location/contains \\

\textbf{Input}: But Mr. Wallace , 45 , a businessman and an investor in partnership for a time with \texttt{<Tail>} Mark Thatcher \texttt{</Tail>} , son of the former British Prime Minister \texttt{<Head>} Margaret Thatcher \texttt{</Head>} , has detractors , including ... . '' \\
\textbf{Output}: /people/person/children \\

\textbf{Input}: Behind the News -- Founded eight years ago in a Silicon Valley garage by two \texttt{<Tail>} Stanford University \texttt{</Tail>} graduate students , \texttt{<Head>} Google \texttt{</Head>} went public two years ago at \$ 85 a share . \\
\textbf{Output}: /business/company/place\_founded \\

\textbf{Input}: A notch or two down-market , the 777 's , 767 's and 757 's are often coveted by corporate titans , among them Larry Page and \texttt{<Tail>} Sergey Brin \texttt{</Tail>} , the co-founders of \texttt{<Head>} Google \texttt{</Head>} , who bought ... . \\
\textbf{Output}: /business/company/founders \newline
/business/company/majorshareholders \\

\textbf{Input}: When he was 32 , \texttt{<Tail>} Bill Hambrecht \texttt{</Tail>} was a co-founder of \texttt{<Head>} Hambrecht \& Quist \texttt{</Head>} , a West Coast investment bank that ... . \\
\textbf{Output}: /business/company/founders
\end{examplebox}

% Query Box
\begin{querybox}
\textbf{Input}: Mr. Bechtolsheim , one of the first investors in \texttt{<Head>} Google \texttt{</Head>} , co-founded Kealia in 2001 with \texttt{<Tail>} David Cheriton \texttt{</Tail>} , a Stanford professor who was another early Google investor .
\end{querybox}

% Output Box
\begin{outputbox}
\hydra: \correct{/business/company/majorshareholders} \\
0-shot: \wrong{/people/person/company}
\end{outputbox}
\caption{\textbf{Example 6:} \hydra\ correctly predicts the ``\textit{Majorshareholders}" relation while 0-shot fails to recognize it.}
\label{fig:relation_examples_3}
\end{figure}
\begin{figure}[ht]
\centering
% Example Box
\begin{examplebox}
\textbf{Input}: ... Mr. Koppel to take another look at a once-unknown man , \texttt{<Head>} Morrie Schwartz \texttt{</Head>} , a \texttt{<Tail>} Brandeis University \texttt{</Tail>} professor who ... . \\
\textbf{Output}: /business/person/company \\

\textbf{Input}: ... as you race away from the pleasant corporate maw of \texttt{<Tail>} Seattle \texttt{</Tail>} , from Starbucks and \texttt{<Head>} Boeing \texttt{</Head>} , Amazon and Microsoft . \\
\textbf{Output}: /business/company/place\_founded \\

\textbf{Input}: ... \texttt{<Head>} Ernest Hemingway \texttt{</Head>} was born in \texttt{<Tail>} Oak Park \texttt{</Tail>} in 1899 and lived here through high school . \\
\textbf{Output}: /place\_of\_birth \newline
/people/person/place\_lived \\

\textbf{Input}: ... Mr. Narayanan 's body will be cremated ... in \texttt{<Tail>} New Delhi \texttt{</Tail>} , near the funeral ground of \texttt{<Head>} Jawaharlal Nehru \texttt{</Head>} , India 's first prime minister ... . \\
\textbf{Output}: /place\_of\_death \newline
/place\_lived \\

\textbf{Input}: The easiest ... way to see \texttt{<Tail>} Philadelphia \texttt{</Tail>} is to stick with the older , central parts of town , emulate \texttt{<Head>} Benjamin Franklin \texttt{</Head>} ... . \\
\textbf{Output}: /place\_of\_death \newline
/place\_lived 
\end{examplebox}

% Query Box
\begin{querybox}
\textbf{Input}: Poe , Evermore A mystery man arrived at \texttt{<Head>} Edgar Allan Poe \texttt{</Head>} 's grave at the Westminster Burial Grounds in \texttt{<Tail>} Baltimore \texttt{</Tail>} on Friday morning , as he has on Poe 's birthday ( Jan. 19 ) every year since 1949 , ... .
\end{querybox}

% Output Box
\begin{outputbox}
\hydra: \correct{/place\_of\_burial, /place\_lived} \\
0-shot: \correct{/place\_of\_burial}
\end{outputbox}
\caption{\textbf{Example 7:} \hydra\ correctly predicts both ``\textit{place\_of\_burial}" and ``\textit{place\_lived}'' while 0-shot partially predicts only one of them.}
\label{fig:relation_examples_4}
\end{figure}

\begin{figure}[ht]
\centering
% Example Box
\begin{examplebox}
\textbf{Input}: ... Mendelssohn , who was born Jewish and converted to Christianity , and \texttt{<Head>} Otto Klemperer \texttt{</Head>} , who converted to Christianity and then back to \texttt{<Tail>} Judaism \texttt{</Tail>} . \\
\textbf{Output}: /religion \\

\textbf{Input}: As the \texttt{<Head>} \texttt{<Tail>} Baltimore \texttt{</Tail>} Orioles \texttt{</Head>} return home ... , Baltimore embraces its rich sports and maritime history . \\
\textbf{Output}: /business/location \\

\textbf{Input}: \texttt{<Head>} Lorenzo Da Ponte \texttt{</Head>} , a Bridge From \texttt{<Tail>} Italy \texttt{</Tail>} to New York '' includes three vocal recitals , beginning tonight with ... . \\
\textbf{Output}: /nationality \\

\textbf{Input}: The Museum of Modern Art 's exhibition of four films starring the \texttt{<Tail>} Italian \texttt{</Tail>} actress \texttt{<Head>} Laura Morante \texttt{</Head>} concludes this weekend with four films , including ... . \\
\textbf{Output}: /ethnicity \\

\textbf{Input}: ... \texttt{<Head>} Madhesi \texttt{</Head>} ethnic group , which by some estimates represents as much as a third of \texttt{<Tail>} Nepal \texttt{</Tail>} 's population of 29 million , has been granted citizenship rights ... . \\
\textbf{Output}: /geographic\_distribution
\end{examplebox}

% Query Box
\begin{querybox}
\textbf{Input}: Among the performances of note : ... the \texttt{<Head>} Italian \texttt{</Head>} dancer ALESSANDRA FERRI gives her final performance with the company on Saturday night , with ROBERTO BOLLE , a guest artist also from \texttt{<Tail>} Italy \texttt{</Tail>} .
\end{querybox}

% Output Box
\begin{outputbox}
Correct: /geographic\_distribution, /nationality

\hydra: \correct{/geographic\_distribution} \\
0-shot: \correct{/nationality}
\end{outputbox}
\caption{\textbf{Example 8:} \hydra\ partially predicts ``\textit{/geographic\_distribution}" while 0-shot partially predicts the other label ``\textit{/nationality}''.}
\label{fig:error_6}
\end{figure}

\begin{figure}[ht]
\centering
% Example Box
\begin{examplebox}
\textbf{Input}: ... owned by a \texttt{<Tail>} Cincinnati \texttt{</Tail>} company , American Financial Group , whose chairman and chief executive officer is Carl H. Lindner III , who is also an owner of the \texttt{<Head>} Cincinnati Reds \texttt{</Head>} . \\
\textbf{Output}: /business/location \\

\textbf{Input}: A biomedical research institute in \texttt{<Tail>} Chengdu \texttt{</Tail>} , \texttt{<Head>} China \texttt{</Head>} , is planning to show true commitment to scientific principles ... . \\
\textbf{Output}: /location/location/contains
/location/country/administrative\_divisions \\

\textbf{Input}: \texttt{<Tail>} Robert Bigelow \texttt{</Tail>} , the founder of \texttt{<Head>} Budget Suites of America \texttt{</Head>} , is likely to push forward ... . \\
\textbf{Output}: /founders \\

\textbf{Input}: A notch or two down-market , the 777 's , 767 's and 757 's are often coveted by corporate titans , among them Larry Page and \texttt{<Tail>} Sergey Brin \texttt{</Tail>} , the co-founders of \texttt{<Head>} Google \texttt{</Head>} , who bought ... . \\
\textbf{Output}: /founders \newline
/majorshareholders \\

\textbf{Input}: The N.F.L. ... is very popular , ... , '' said Tony Ponturo , vice president for global media and sports marketing at \texttt{<Head>} Anheuser-Busch \texttt{</Head>} in \texttt{<Tail>} St. Louis \texttt{</Tail>} , ... . '' \\
\textbf{Output}: /place\_founded
\end{examplebox}

% Query Box
\begin{querybox}
\textbf{Input}: \texttt{<Head>} Nestlé \texttt{</Head>} , based in \texttt{<Tail>} Vevey \texttt{</Tail>} , Switzerland , ... .
\end{querybox}

% Output Box
\begin{outputbox}
Correct: /business/location \\
\hydra: \wrong{/place\_founded} \\
0-shot: \wrong{/place\_founded}
\end{outputbox}
\caption{\textbf{Example 9:} Both \hydra\ and 0-shot misclassify the \textit{``/business/location"} label as \textit{``/place\_founded"}}
\label{fig:error_1}
\end{figure}

\begin{figure}[ht]
\centering
% Example Box
\begin{examplebox}
\textbf{Input}: But Mr. Wallace ... in partnership for a time with \texttt{<Tail>} Mark Thatcher \texttt{</Tail>} , son of the former British Prime Minister \texttt{<Head>} Margaret Thatcher \texttt{</Head>} , has detractors ... . \\
\textbf{Output}: /people/person/children \\

\textbf{Input}: Because the Newbery is open only to American citizens or residents , one enormously popular writer who is n't in the running is \texttt{<Head>} Cornelia Funke \texttt{</Head>} , who lives in \texttt{<Tail>} Germany \texttt{</Tail>} and whose books appear here in translation . \\
\textbf{Output}: /people/person/nationality \\

\textbf{Input}: \texttt{<Head>} Keith Jarrett \texttt{</Head>} lives on the New Jersey side of the Pennsylvania border , within an hour 's drive of his childhood home of \texttt{<Tail>} Allentown \texttt{</Tail>} , Pa. . \\
Output: /people/person/place\_of\_birth \newline
/people/person/place\_lived  \\

\textbf{Input}: ... , Pollan finds his hero in \texttt{<Head>} Joel Salatin \texttt{</Head>} , an '' alternative '' farmer in \texttt{<Tail>} Virginia \texttt{</Tail>} ... . \\
\textbf{Output}: /people/person/place\_lived \\

\textbf{Input}: Charles G. Taylor , the former president of Liberia ... arrived in handcuffs in the \texttt{<Tail>} Netherlands \texttt{</Tail>} on Tuesday , and was immediately taken to the jail near \texttt{<Head>} The Hague \texttt{</Head>} ... . \\
\textbf{Output}: /administrative\_division/country
\end{examplebox}

% Query Box
\begin{querybox}
\textbf{Input}: Mr. Meshal lives in exile in \texttt{<Head>} Damascus \texttt{</Head>} , \texttt{<Tail>} Syria \texttt{</Tail>} , where the government has rebuffed all Western requests to close his office .
\end{querybox}

% Output Box
\begin{outputbox}
Correct: /administrative\_division/country \\
\hydra: \wrong{/location/location/contains} \\
0-shot: \wrong{/location/location/contains}
\end{outputbox}
\caption{\textbf{Example 10:} Both \hydra\ and 0-shot misclassify the \textit{``/administrative\_division/country"} label as \textit{``/location/location/contains"}}
\label{fig:error_2}
\end{figure}

\begin{figure}[ht]
\centering
% Example Box
\begin{examplebox}
\textbf{Input}: ... the epic poem '' Paterson '' by \texttt{<Head>} William Carlos Williams \texttt{</Head>} , a native of \texttt{<Tail>} Rutherford \texttt{</Tail>} . \\
\textbf{Output}: /people/person/place\_of\_birth \\

\textbf{Input}: It goes on to list notable \texttt{<Tail>} Mississippi \texttt{</Tail>} writers including William Faulkner , \texttt{<Head>} Richard Wright \texttt{</Head>} , ... . \\
\textbf{Output}: /people/person/place\_lived \\

\textbf{Input}: ... , including Giocangga , the founder of the \texttt{<Head>} Manchu \texttt{</Head>} dynasty in \texttt{<Tail>} China \texttt{</Tail>} , and Niall of the Nine Hostages , ... . '' \\
\textbf{Output}: /people/ethnicity/geographic\_distribution \\

\textbf{Input}: \texttt{<Head>} Anthony Trollope \texttt{</Head>} , the brilliant depicter of the 19th-century social strata in \texttt{<Tail>} England \texttt{</Tail>} , ... . \\
\textbf{Output}: /people/person/nationality \\

\textbf{Input}: ... , Mr. Delli Colli worked with generations of \texttt{<Tail>} Italian \texttt{</Tail>} directors , including \texttt{<Head>} Pier Paolo Pasolini \texttt{</Head>} , ... . \\
\textbf{Output}: /people/person/ethnicity
\end{examplebox}

% Query Box
\begin{querybox}
\textbf{Input}: Lukacs , a distinguished historian of 20th-century Europe , makes very large claims for his subject in '' George Kennan '' : He was '' a better writer and a better thinker '' than \texttt{<Head>} Henry Adams \texttt{</Head>} ; he was '' the best and finest \texttt{<Tail>} American \texttt{</Tail>} writer about Europe '' in the interwar years , better than Hemingway .
\end{querybox}

% Output Box
\begin{outputbox}
Correct: /people/person/ethnicity \\
\hydra: \wrong{/people/person/nationality} \\
0-shot: \wrong{/people/person/nationality}
\end{outputbox}
\caption{\textbf{Example 11:} Both \hydra\ and 0-shot misclassify the \textit{``/people/person/ethnicity"} label as \textit{``/people/person/nationality"}}
\label{fig:error_3}
\end{figure}

\begin{figure}[ht]
\centering
% Example Box
\begin{examplebox}
\textbf{Input}: Halliburton operates in Iran through a unit based in nearby \texttt{<Head>} Dubai \texttt{</Head>} , \texttt{<Tail>} United Arab Emirates \texttt{</Tail>} , ... . \\
\textbf{Output}: /administrative\_division/country \\

\textbf{Input}: ... when the State Department sponsored its tour to Damascus , Homs and \texttt{<Tail>} Lattakia \texttt{</Tail>} , \texttt{<Head>} Syria \texttt{</Head>} . \\
\textbf{Output}: /location/location/contains \newline
/country/administrative\_divisions \\

\textbf{Input}: That is one reason that \texttt{<Head>} Hunan \texttt{</Head>} 's fast-growing provincial capital , \texttt{<Tail>} Changsha \texttt{</Tail>} , is beginning to ... . \\
\textbf{Output}: /location/location/contains \newline
/region/capital \\

\textbf{Input}: It has outfitted the World Financial Center and the new Bloomberg L.P. headquarters in New York as well as the \texttt{<Tail>} Cheung Kong Center \texttt{</Tail>} in \texttt{<Head>} Hong Kong \texttt{</Head>} ... . \\
\textbf{Output}: /location/location/contains \\

\textbf{Input}: There is no shortage of cruises that stop in either \texttt{<Tail>} Tallinn \texttt{</Tail>} , \texttt{<Head>} Estonia \texttt{</Head>} , or Riga , Latvia , or both . \\
\textbf{Output}: /location/location/contains \newline
/location/country/capital
\end{examplebox}

% Query Box
\begin{querybox}
\textbf{Input}: Last year , it opened offices in Warsaw and \texttt{<Tail>} Bucharest \texttt{</Tail>} , the capital of \texttt{<Head>} Romania \texttt{</Head>} .
\end{querybox}

% Output Box
\begin{outputbox}
Correct: /country/capital, /country/administrative\_divisions, /location/location/contains \\
\hydra: \correct{/location/location/contains, 
/country/capital} \\
0-shot: \correct{/country/capital}
\end{outputbox}
\caption{\textbf{Example 12:} \hydra\ misses the \textit{``/country/administrative\_divisions"} label while 0-shot misses both \textit{``/country/administrative\_divisions"} and /location/location/contains.}
\label{fig:error_4}
\end{figure}

\begin{figure}[ht]
\centering
% Example Box
\begin{examplebox}
\textbf{Input}: Mr. Schwartz , 32 , teaches English ... as a \texttt{<Tail>} New York City \texttt{</Tail>} teaching fellow at Intermediate School 349 in \texttt{<Head>} Bushwick \texttt{</Head>} , Brooklyn . \\
\textbf{Output}: /neighborhood\_of \\

\textbf{Input}: \texttt{<Tail>} Nyack \texttt{</Tail>} 's mayor , John Shields , said ... to locate a theater company in the area if efforts to save the \texttt{<Head>} Helen Hayes \texttt{</Head>} theater fail . \\
\textbf{Output}: /place\_of\_death \\

\textbf{Input}: ``..." said Mr. Jordan , who just signed on with \texttt{<Head>} Mark Warner \texttt{</Head>} , a Democrat and the former governor of \texttt{<Tail>} Virginia \texttt{</Tail>} who is considering a run for president . \\
\textbf{Output}: /place\_lived \\

\textbf{Input}: \texttt{<Head>} John David Booty \texttt{</Head>} 's decision to skip his senior year at ... in \texttt{<Tail>} Shreveport \texttt{</Tail>} did not ... . \\
\textbf{Output}: /place\_of\_birth,
/place\_lived \\

\textbf{Input}: Jack Emmert had already earned his master 's degree ... to become creative director at \texttt{<Head>} Cryptic Studios \texttt{</Head>} , a game company based in \texttt{<Tail>} Los Gatos \texttt{</Tail>} , Calif. , where he ... . \\
\textbf{Output}: /place\_founded

\end{examplebox}

% Query Box
\begin{querybox}
\textbf{Input}: But before Mr. Joffe , ... , accepted a job , a friend suggested he check out a \texttt{<Tail>} San Francisco \texttt{</Tail>} start-up , \texttt{<Head>} Powerset \texttt{</Head>} , which is trying to build a rival search engine .
\end{querybox}

% Output Box
\begin{outputbox}
Correct: /business/location, /place\_founded \\
\hydra: \correct{/place\_founded}\\
0-shot: \correct{/place\_founded}
\end{outputbox}
\caption{\textbf{Example 13:} Both \hydra\ and 0-shot GPT-4o miss the obvious label \textit{``/business/location"} while correctly predict a more nuanced label \textit{``/place\_founded"}.}
\label{fig:error_5}
\end{figure}

% \begin{figure}[ht]
% \includegraphics[width=0.95\columnwidth, height=8cm]{conf_all.png}
%     \caption{Confusion analysis of \hydra\ vs the baselines (aggregated over all relations)}
%     \label{fig:conf_all}
% \end{figure} 

% \begin{table}[ht]
% %\vspace*{-0.5ex}
% \centering
% \small{\begin{tabular}{@{}lrrrl@{}}
%   {} & \textbf{Baseline correct} &\textbf{Baseline Wrong} \\ 
%  \hline
%  \textbf{\hydra\ correct} & 312 & 83\\
%  \textbf{\hydra\ wrong} & 30  & 267\\
% \end{tabular}}
% \caption{Confusion analysis of \hydra\ (5-shot GPT-4o) vs 0-shot GPT-4o for English.}
% \label{table:conf1}
% %\vspace*{-1ex}
% \end{table}
% \begin{table}[ht]
% %\vspace*{-0.5ex}
% \centering
% \small{\begin{tabular}{@{}lrrrl@{}}
%   {} & \textbf{Baseline correct} &\textbf{Baseline Wrong} \\ 
%  \hline
%  \textbf{\hydra\ correct} & 172 & 223\\
%  \textbf{\hydra\ wrong} & 25 & 272\\
% \end{tabular}}
% \caption{Confusion analysis of \hydra\ (5-shot GPT-4o) vs PARE for English.}
% \label{table:conf2}
% %\vspace*{-1ex}
% \end{table}

\begin{figure*}[t]
    \centering
    \captionsetup[subfigure]{labelformat=simple, labelsep=space, font=small}
    % 7th row
    \begin{subfigure}[t]{0.48\linewidth}
        \includegraphics[width=\linewidth, valign=t]{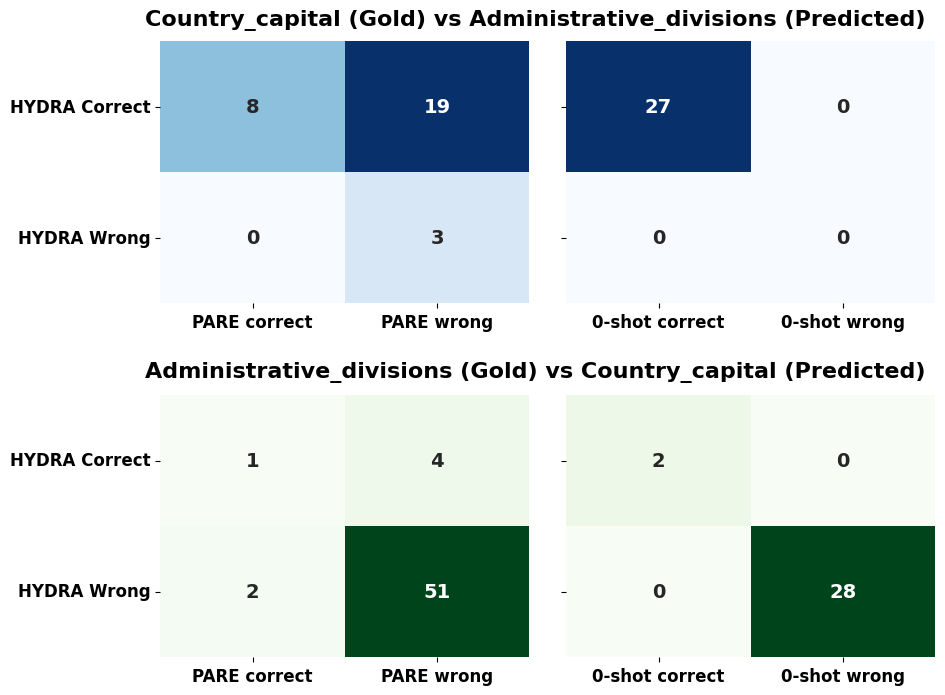}
        \caption{Capital vs Administrative\_divisions}
        \label{fig:conf_m}
    \end{subfigure}
    \hfill
    \begin{subfigure}[t]{0.48\linewidth}
        \includegraphics[width=\linewidth, valign=t]{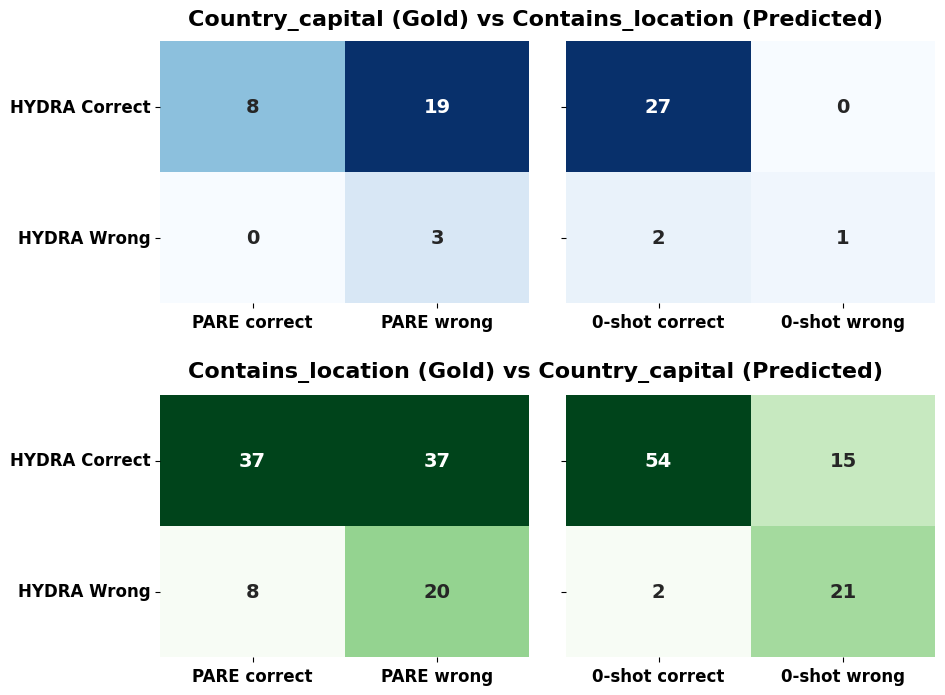}
        \caption{Capital vs Contains\_location}
        \label{fig:conf_n}
    \end{subfigure}
    \begin{subfigure}[t]{0.48\linewidth}
        \includegraphics[width=\linewidth, valign=t]{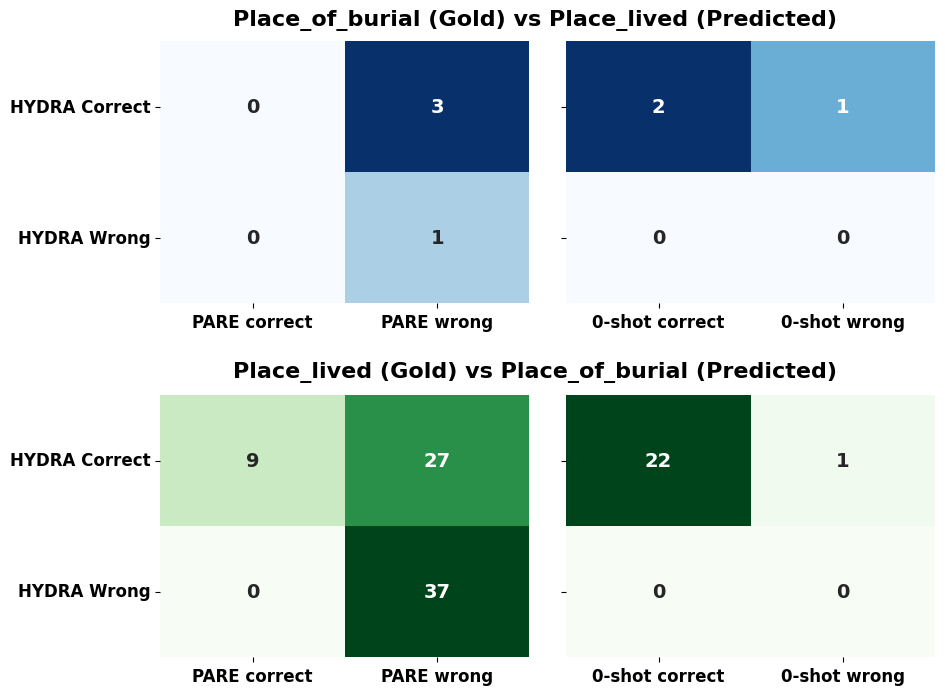}
        \caption{Place of Burial vs Place lived}
        \label{fig:conf_o}
    \end{subfigure}
    \hfill
    \begin{subfigure}[t]{0.48\linewidth}
        \includegraphics[width=\linewidth, valign=t]{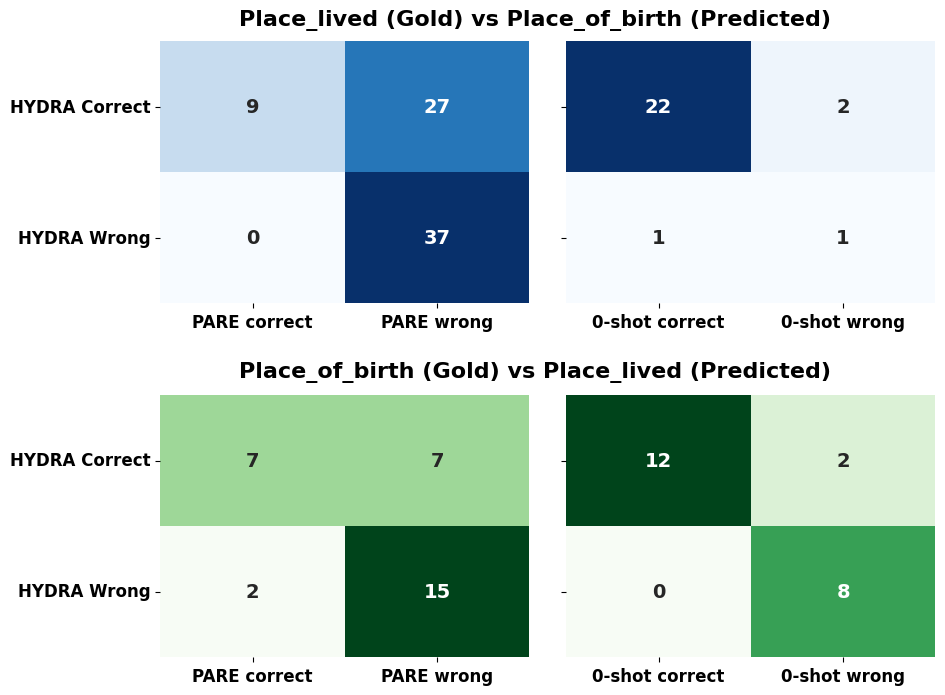}
        \caption{Place of birth vs Place lived}
        \label{fig:conf_p}
    \end{subfigure}
    \caption{Confusion matrices analyzing \hydra\ vs the baselines across relation types}
    \label{fig:relation_matrices_c}
\end{figure*}

\subsection{Recall$@$5 for Supervised DSRE models}
\label{subsec: recall5}
We present Recall$@$5 for both \parex\ and \cilx\ models in table \ref{table:recall-5} for English and other languages. 
\begin{table}[ht]
%\vspace*{-0.5ex}
\centering
\small{\begin{tabular}{@{}lrrrl@{}}
  \textbf{Language} &\textbf{\pare\ } & \textbf{\cil\ } \\ 
 \hline
 English & \textbf{84} & 82\\
 \hdashline
 Oriya & \textbf{75} & 72\\
 Santhali & \textbf{72} & 64\\
 Manipuri & 69 & 69\\
 Tulu & 69 & \textbf{71}\\
 \hline
 Avg.* & \textbf{71} & 69
\end{tabular}}
\caption{Recall$@$5 scores for \pare\ and \cil\ models on all languages. *Averaged over non-English languages}
\label{table:recall-5}
%\vspace*{-1ex}
\end{table}
\subsection{Additional supervised DSRE baselines}
We include a comparison of \pare\ and \cil\ with HiCLRE and HFMRE baselines in Table \ref{table:sup}.
\begin{table}[ht]
%\vspace*{-0.5ex}
\centering
\small{\begin{tabular}{@{}lrrrl@{}}
  \textbf{Baseline} &\textbf{Micro F1} & \textbf{Macro F1} \\ 
 \hline
 HiCLRE & 31 & 18\\
 HFMRE & 32 & 18\\
 \hdashline
 PARE & 42 & 31\\
 CIL & \textbf{43} & \textbf{32}\\
\end{tabular}}
\caption{Comparison of additional English baselines with PARE and CIL.}
\label{table:sup}
%\vspace*{-1ex}
\end{table}

\subsection{Scalability w.r.t. No. of candidates k}
\label{subsec:scale}
\begin{figure*}[ht]
    \centering
    \captionsetup[subfigure]{labelformat=simple, labelsep=space, font=small}
    % 7th row
    \begin{subfigure}[t]{0.48\linewidth}
        \includegraphics[width=0.95\linewidth, valign=t]{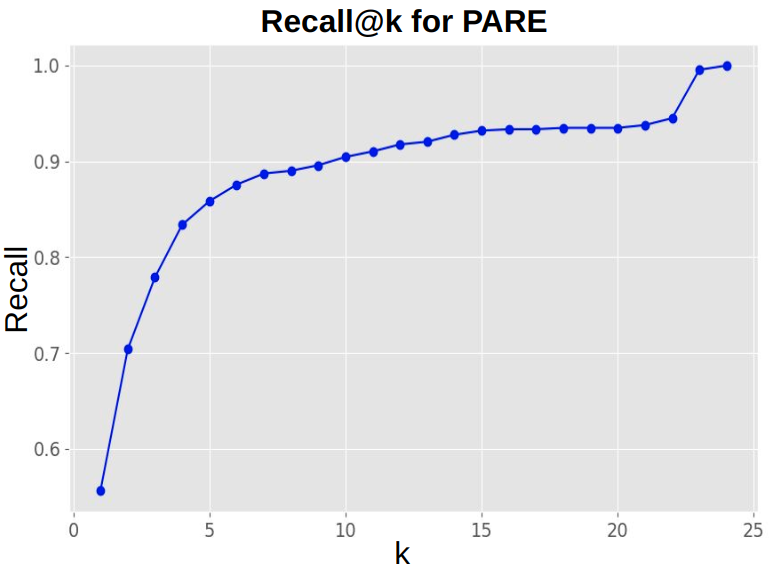}
        \caption{\pare\ Recall\texttt{@}k v/s k}
        \label{fig:rec_k}
    \end{subfigure}
    \hfill
    \begin{subfigure}[t]{0.48\linewidth}
        \includegraphics[width=0.95\linewidth, valign=t]{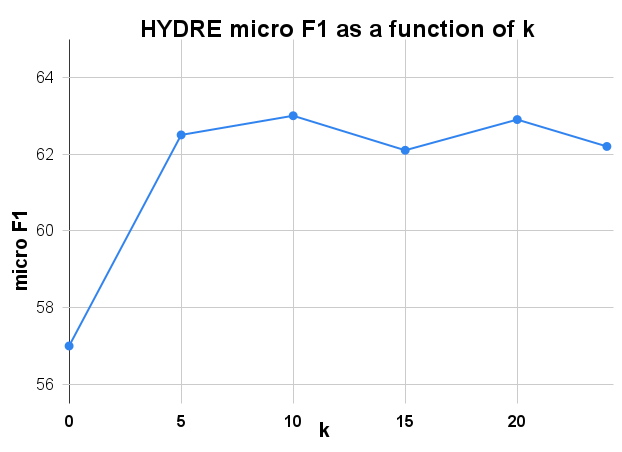}
        \caption{\hydra\ (GPT-4o) micro-F1 v/s k}
        \label{fig:f1_k}
    \end{subfigure}
    \caption{Analysis of PARE's Recall\texttt{@}k and \hydra\ (GPT-4o) downstream performance on NYT-10m English dev set}
    \label{fig:reck}
\end{figure*}

We analyze how PARE's Recall\text@k and the \hydra\ performance vary as a function of $k$ (no. of candidate relations) on the NYT-10m English dev set. Results are shown in Figure \ref{fig:reck}. While the recall\text@k keeps on increasing even beyond k=5, the downstream F1 performance saturates at k = 5 showing that increasing k may not only increase LLM's prompt length but also confuse the LLM since the number of candidates is too large to filter the correct relation(s). Therefore, we fix k=5 consistently for all our experiments.
% \begin{figure}[ht]
% \includegraphics[width=0.4\textwidth, height=10cm]{Recall_k_F1_k.jpeg}
%     \caption{Scalability of \pare's Recall\text@k (top) \& downstream F1 score (bottom) as a function of k (No. of candidates)}
%     \label{fig:reck}
% \end{figure} 

\subsection{Detailed Language-wise Ablations}
Results are presented in tables \ref{table:ablat4}, \ref{table:ablat5}, \ref{table:ablat5-qwen} and \ref{table:ablat3} for each of the 4 LLMs.
\begin{table*}[ht]
%\vspace*{-0.5ex}
\centering
\small{\begin{tabular}{@{}lrrrrrrrl@{}}
  \textbf{Ablation variant} &\textbf{Ory} & \textbf{Sat}  & \textbf{Mni} & \textbf{Tcy} & \textbf{mi.} & \textbf{mu.} \\ 
 \hline
 \hydra\ & 38/22    & 24/16 & \textbf{25}/\textbf{13} & \textbf{42}/\textbf{28} & 32 & \textbf{20}  \\
 w. sem. sim. & 34/24 & 21/16 & 14/10 & 35/22   & 26 & 18  \\
 w/o \pare\ (Random) &  33/22 & 24/14 & 17/10 & 30/18 & 26 &    16 \\
 w/o stage 1 & \textbf{39}/\textbf{26} & \textbf{33}/\textbf{20} & 23/10 & 38/25 & \textbf{33} & \textbf{20} \\
 w/o ICL & 32/21    & 28/19 & 25/16 & 30/19 & 29    & 19
 % w/o ontology & 30/20 & 24/17 & 18/14 & 31/21 & 26 & 18  \\
\end{tabular}}
\caption{Average F1 scores for \textbf{Llama3.1-8b} over our target languages for different ablation variants of our few-shot approach in \textit{translate-train} setting}
\label{table:ablat4}
%\vspace*{-1ex}
\end{table*}
\begin{table*}[ht]
%\vspace*{-0.5ex}
\centering
\small{\begin{tabular}{@{}lrrrrrrrl@{}}
  \textbf{Ablation variant} &\textbf{Ory} & \textbf{Sat}  & \textbf{Mni} & \textbf{Tcy} & \textbf{mi.} & \textbf{mu.} \\ 
 \hline
 \hydra\ & \textbf{56}/39    & \textbf{39}/\textbf{22} & \textbf{35}/\textbf{16} & \textbf{51}/\textbf{36} & \textbf{45} & \textbf{28} \\
 w. sem. sim. & 54/\textbf{40} & 38/\textbf{22} & 33/14 & 49/32   & 44 & 27  \\
 w/o \pare\ (Random) &  53/34 & 31/15 & 21/8 & 47/28 & 38   & 21 \\
 % w/o ontology & 57/39 & 37/22 & 31/11 & 48/31 & 43 & 26 \\
 w/o stage 1 & 54/35 & 35/21 & 33/13 & 48/30 & 43 & 25 \\
 w/o ICL & 50/30    & 27/15 & 10/6  & 44/22 & 33    & 18
\end{tabular}}
\caption{Average F1 scores for \textbf{Llama3.1-8b-FT$_{X}$} over our target languages for different ablation variants of our few-shot approach in \textit{translate-train} setting}
\label{table:ablat5}
%\vspace*{-1ex}
\end{table*}

\begin{table*}[ht]
%\vspace*{-0.5ex}
\centering
\small{\begin{tabular}{@{}lrrrrrrrl@{}}
  \textbf{Ablation variant} &\textbf{Ory} & \textbf{Sat}  & \textbf{Mni} & \textbf{Tcy} & \textbf{mi.} & \textbf{mu.} \\ 
 \hline
 \hydra\ &  \textbf{56}/53   & 23/12 & 14/6 & 57/52 & \textbf{38} & \textbf{31} \\
 w. sem. sim. & 55/53 & \textbf{24}/\textbf{15} & 15/5 & 55/52   & 37 & \textbf{31}  \\
 w/o \pare\ (Random) & 55/\textbf{55} & 19/11 & 15/5 & \textbf{58}/\textbf{53} & 37 & \textbf{31}  \\
 w/o stage 1 & 51/52 & 17/9 & \textbf{16}/\textbf{7} & 52/50 & 34 & 30  \\
 w/o ICL &  54/36   & 15/8 & 8/4  & 53/33 & 33    & 20  \\
\end{tabular}}
\caption{Average F1 scores for \textbf{Qwen3-235B-A22B} over our target languages for different ablation variants of our few-shot approach in \textit{translate-train} setting}
\label{table:ablat5-qwen}
%\vspace*{-1ex}
\end{table*}
\begin{table*}[ht]
%\vspace*{-0.5ex}
\centering
\small{\begin{tabular}{@{}lrrrrrrrl@{}}
  \textbf{Ablation variant} &\textbf{Ory} & \textbf{Sat}  & \textbf{Mni} & \textbf{Tcy} & \textbf{mi.} & \textbf{mu.}\\ 
 \hline
 \hydra\ & \textbf{58}/56  & 17/11  & 20/\textbf{15} & \textbf{57}/55  & 38 & 34  \\
 w. sem. sim. & 57/\textbf{57} & 18/\textbf{12} & \textbf{22}/\textbf{15}   & 56/52 & 38 & 34  \\
 w/o \pare\ (Random) & 55/55 & 15/8 & 16/10 & \textbf{57}/55 & 36 & 32  \\
 w/o stage 1 & 57/56 & \textbf{20}/11 & 20/\textbf{15} & \textbf{57}/\textbf{57} & \textbf{39} & \textbf{35}\\
 w/o ICL & 54/38    & 12/6 & 12/7 & 48/32 & 32 & 21  \\
 % w/o ontology & 55/52 & 20/11 & 17/10 & 56/53 & 37  & 32  \\
\end{tabular}}
\caption{Average F1 scores for \textbf{GPT-4o} over our target languages for different ablation variants of our few-shot approach in \textit{translate-train} setting}
\label{table:ablat3}
%\vspace*{-1ex}
\end{table*}
\subsection{Dataset statistics}
\label{subsec: testdata}
\subsubsection{Test data split}
We construct the test split from NYT-10m (Gao et al. 2021) using stratified sampling, ensuring a minimum of 30 instances per relation, except when a relation contains fewer than 30 instances in total. This results in 538 sentences, with the distribution of relation counts shown in Table~\ref{table:labstat}. Since NYT-10m is a multi-label dataset, the total number of relation instances in the test split exceeds the number of sentences, amounting to 722.
\subsubsection{Training data}
We take the original training data from NYT-10m (Gao et al. 2021) and ensure that ``NA'' bags do not exceed 10\% of total bags to avoid model overfit on ``NA'' label. This leads to a total number of 41624 training bags. 
\begin{table}[ht]
%\vspace*{-0.5ex}
\centering
\small{\begin{tabular}{@{}lrrl@{}}
  \textbf{Relation} &\textbf{Count} \\ 
 \hline
 /people/person/place\_lived & 73\\
 /people/person/nationality & 34\\
 /business/person/company & 34 \\
 /people/person/place\_of\_birth & 31 \\
 /location/location/contains & 102 \\
 /location/country/administrative\_divisions & 58 \\
 /business/location & 33 \\
 /location/administrative\_division/country & 31 \\
 /business/company/advisors & 37 \\
 /business/company/founders & 31 \\
 /business/company/majorshareholders & 4 \\
 /location/neighborhood/neighborhood\_of & 30 \\
 /location/country/capital & 30 \\
 /film/film/featured\_film\_locations & 1 \\
 /location/us\_county/county\_seat & 6 \\
 /people/person/children & 30 \\
 /people/deceasedperson/place\_of\_death & 30 \\
 /people/deceasedperson/place\_of\_burial & 4 \\
 /people/ethnicity/geographic\_distribution & 30 \\
 /location/region/capital & 12 \\
 /business/company/place\_founded & 4 \\
 /people/person/religion & 21 \\
 /time/event/locations & 3 \\
 /people/person/ethnicity & 23 \\
  NA & 30\\
 \hline
 Total & 538
\end{tabular}}
\caption{Label-wise statistics of our evaluation data. Total number of sentences (538) in our test split is different from total number of labels (722) due to multi-label characteristics of NYT-10m dataset.}
\label{table:labstat}
%\vspace*{-1ex}
\end{table}
\subsection{Data annotation for Indic languages} 
\label{subsec:indictest}
\subsubsection{Annotator details}
We conduct human verification for four Indic languages using native speakers—either students or IT professionals—who were proficient in reading and typing in their respective scripts. Each annotator was compensated approximately \$60 for verifying translations of 538 sentences. Prior to annotation, the speakers were informed that the task was intended solely for research purposes and posed no risk to them.

Each annotator was presented with the following questionnaire, with binary (YES/NO) responses and rectifications requested in case of a NO:

\begin{enumerate}
\item \textbf{Q1.} Is the translation of the given English sentence correct?
\item \textbf{Q2.} Is the \textit{head entity} correctly translated into your native language?
\item \textbf{Q3.} Is the \textit{head entity} correctly projected in your native language?
\item \textbf{Q4.} Is the \textit{tail entity} correctly translated into your native language?
\item \textbf{Q5.} Is the \textit{tail entity} correctly projected in your native language?
\end{enumerate}
\subsubsection{Quality Assessment and Interannotator Agreement}
We first assess the quality of the system-generated translations presented to annotators. Native speakers across all languages found the translations to be generally decent—likely due to high-quality output from IndicTrans2 (and Google Translate in the case of Tulu).

To quantify this, Table \ref{table:sys-trans} reports:
\begin{enumerate}
    \item The percentage of translations that required no human correction
    \item The character-level F1 score (Char-F1) between the original system-generated translation and the human-corrected version (for cases requiring rectification).
\end{enumerate}

\begin{table}[ht]
%\vspace*{-0.5ex}
\centering
\small{\begin{tabular}{@{}lrrrl@{}}
  \textbf{Language} & \textbf{No Rectification} & \textbf{Char-F1 Match} \\
  & \textbf{Needed (\%)} & \textbf{(if rectified)} \\
  \hline
Oriya & 69 & 92 \\
Santhali & 72 & 88 \\
Manipuri & 83 & 96 \\
Tulu & 73 & 95 \\
\hline
Average & 74 & 93 \\
\end{tabular}}
\caption{Percentage of system translations requiring no rectification, and character-level F1 match for rectified translations.}
\label{table:sys-trans}
%\vspace*{-1ex}
\end{table}
To further evaluate annotation reliability, we conducted an inter-annotator agreement study. A second native speaker independently judged the quality of translations for 100 randomly sampled sentences in each language. Agreement is reported as the percentage of samples where the second speaker’s judgment matched that of the first. Results are shown in Table~\ref{table:inter-ann}.
\begin{table}[ht]
%\vspace*{-0.5ex}
\centering
\small{\begin{tabular}{@{}lrrrrl@{}}
  \textbf{Language} & \textbf{Translation (\%)} & \textbf{Head entity} & \textbf{Tail entity} \\
  & & \textbf{projection} & \textbf{projection} \\
  \hline
Oriya & 92 & 92 & 92 \\
Santhali & 89 & 84 & 87 \\
Manipuri & 85 & 98 & 100 \\
Tulu & 96 & 92 & 88 \\
\hline
Average & 91 & 92 & 92 \\
\end{tabular}}
\caption{Inter-annotator agreement for 100 samples in each language}
\label{table:inter-ann}
%\vspace*{-1ex}
\end{table}

In summary, on average, 74\% of system-generated translations were accepted as correct by native speakers, and for the remaining, the human-corrected outputs had a 93\% Char-F1 match with the original translations. Inter-annotator agreement for human-corrected outputs averaged 91\%, indicating strong consistency and translation quality across languages.

\subsection{Semantic retrieval hurts performance for low-resource languages}
\label{subsec: sem-x}
We try to use both off-the-shelf retriever BGE-m3 and a fine-tuned retriever (\cilx's sentence encoder) for \hydra\ in translate-train setting. We observe (Table \ref{table:ablat6}) that though the fine-tuned retriever's performance is significantly better than BGE-m3, they both are worse compare to the variant of \hydra\ that only uses \parex's confidence for candidate scoring (results in table). This analysis suggest that semantic similarity is not useful in translate-train settings for our target languages.

\begin{table}[ht]
%\vspace*{-0.5ex}
\centering
\small{\begin{tabular}{@{}lrrrrl@{}}
  \textbf{Ablation variant} &\textbf{Llama3.1} & \textbf{Llama3.1-ft}  & \textbf{GPT-4o} \\ 
 \hline
 % ours & 26 & 43 & 39  \\
 \hydra\ (using only  & \textbf{32} & \textbf{45} &  38\\
\parex\ confidence) & & & & \\
 sem. sim. (BGE-M3) & 24 & 42 &  36\\
 sem. sim. (\cilx)  &  26 & 44 & \textbf{39}\\
\end{tabular}}
\caption{Average F1 scores over our target languages for \hydra\ (using only \parex\ confidence) and sem. sim. variants (both off-the-shelf and fine-tuned) in \textit{translate-train} setting. }
\label{table:ablat6}
%\vspace*{-1ex}
\end{table}
\subsection{Additional ablation to justify aggregate confidence for sentence selection}
\label{subsec: agg_score}
\hydra\ selects the representative sentence based on aggregated confidence over all labels of the bag. This helps surface sentences with stronger overall evidence, and not just noisy single-label confidence.
We justify our design choice with an additional ablation \textit{Candidate-only scoring} in which we use only candidate label's confidence score for stage 2 retrieval instead of aggregate score over all bag labels. Results are shown in Table \ref{table:ablat3}. 
% \begin{itemize}
%     \item \textit{Candidate-only scoring} (use only candidate label's confidence) 
%     \item \textit{Random} sentence selection
% \end{itemize}
\begin{table}[ht]
%\vspace*{-0.5ex}
\centering
\small{\begin{tabular}{@{}lrrrrl@{}}
  \textbf{Ablation variant} & \textbf{GPT-4o} & \textbf{Qwen3} & \textbf{Llama3.1} \\ 
 \hline
\hydra\ (aggregate scoring) &   \textbf{63}/60 & \textbf{63}/\textbf{62} & \textbf{52}/\textbf{47} \\
Candidate-only scoring & \textbf{63}/\textbf{62} & 62/60 & 47/35 \\
% Random sentence selection & 53/43 & 62/61 & 32/17 \\
\end{tabular}}
\caption{Ablations of confidence aggregation step for sentence selection in stage 2.}
\label{table:ablat7}
%\vspace*{-1ex}
\end{table}
While candidate-only scoring performs comparably for GPT-4o, aggregate scoring yields substantial gains for smaller LLMs like Llama and Qwen3, validating its broader effectiveness.

\subsection{Additional results and ablations on Wiki-20m}
\label{subsec: wiki}
We expand our evaluation on another English dataset Wiki-20m (Gao et al., 2021), containing 2386 manually annotated test sentences and 81 relations (including ``NA''). Fig. \ref{table:wiki} presents the results and ablations for GPT-4o and Llama-3.1-8b-Instruct models. We observe that \hydra\ achieves huge gains compared to 0-shot and ablations, beating it's closest competitor by 10 and 20 micro F1 points respectively for GPT-4o and Llama-3.1, showing that our approach also scales well when the number of relations in the ontology is large (80 in this case). 

\begin{table}[ht]
%\vspace*{-0.5ex}
\centering
\small{\begin{tabular}{@{}lrrrrl@{}}
  \textbf{Ablation} &\textbf{Llama3.1}   &  \textbf{GPT-4o} \\ 
 \hline
 0-shot  & 11/16 & 55/48 \\
 \hydra  & \textbf{54/53} & \textbf{67/63} \\
 \hdashline
 \textit{Ablations} & \\
 w/o semantic similarity  & 25/29 & 34/36 \\
 w/o PARE confidence & 21/16 & 40/36  \\
 w/o both (Random) & 34/34 & 57/54 \\
 w/o stage 2  & 39/39 & 38/37 \\
 w/o ICL & 3/4 & 12/4
 % w/o ontology  & 54/48 & 62/59 \\
\end{tabular}}
\caption{English F1 (micro/macro) scores on Wiki-20m evaluation set.}
\label{table:wiki}
%\vspace*{-1ex}
\end{table}
\end{document}